\documentclass[conference]{IEEEtran}
\ifCLASSINFOpdf
 \usepackage[pdftex]{graphicx}
\else
\usepackage[dvips]{graphicx}
\fi
\usepackage[cmex10]{amsmath}
\usepackage{stfloats}

\usepackage{booktabs}	

 \usepackage[bookmarks=false,
						plainpages=false,																
						pdfpagelabels,
						linktocpage=true,			
						colorlinks=true,			
						linkcolor=blue,				
						citecolor=red,				
						urlcolor=magenta,			
						hypertexnames=false]
					{hyperref}

\newcounter{MYtempeqncnt}
\newcommand*\rfrac[2]{{}^{#1}\!/_{#2}}

\newcommand{\mysmallarraydecl}{\renewcommand{%
\IEEEeqnarraymathstyle}{\scriptscriptstyle}%
\renewcommand{\IEEEeqnarraytextstyle}{\scriptsize}%
\renewcommand{\baselinestretch}{1.1}%
\settowidth{\normalbaselineskip}{\scriptsize
\hspace{\baselinestretch\baselineskip}}%
\setlength{\baselineskip}{\normalbaselineskip}%
\setlength{\jot}{0.25\normalbaselineskip}%
\setlength{\arraycolsep}{2pt}}

\begin{document}

\onecolumn

{\textcircled{c}} 2017 IEEE. Personal use of this material is permitted. Permission from IEEE must be obtained for all
other uses, in any current or future media, including reprinting/republishing this material for advertising
or promotional purposes, creating new collective works, for resale or redistribution to servers or lists, or
reuse of any copyrighted component of this work in other works. \\

Conference: IEEE OCEANS 2014 - TAIPEI , 7-10 April 2014 \\

DOI: 10.1109/OCEANS-TAIPEI.2014.6964475 \\

URL: http://ieeexplore.ieee.org/document/6964475/

\twocolumn

\title{Model Identification and Controller Parameter Optimization for an Autopilot Design for Autonomous Underwater Vehicles}
\author{
\IEEEauthorblockN{Ralf Taubert, Mike Eichhorn, Christoph Ament}
\IEEEauthorblockA{Inst. for Automation and Systems Engineering Ilmenau\\
									University of Technology\\
									98693 Ilmenau, Germany\\
									Email: \href{mailto:\%7bralf.taubert,\%20mike.eichhorn,\%20christoph.ament\%7d@tu-ilmenau.de}{\{ralf.taubert, mike.eichhorn, christoph.ament\}}\\
												 \href{mailto:\%7bralf.taubert,\%20mike.eichhorn,\%20christoph.ament\%7d@tu-ilmenau.de}{@tu-ilmenau.de}} 
\and
\IEEEauthorblockN{Marco Jacobi, Divas Karimanzira, Torsten Pfuetzenreuter}
\IEEEauthorblockA{Advanced System Technology (AST)\\
									Branch of Fraunhofer IOSB\\
									98693 Ilmenau, Germany\\
									Email: \href{mailto:\%7bmarco.jacobi,\%20divas.karimanzira,\%20torsten.pfuetzenreuter\%7d@iosb-ast.fraunhofer.de}{\{marco.jacobi, divas.karimanzira, torsten.pfuetzenreuter\}}\\
												 \href{mailto:\%7bmarco.jacobi,\%20divas.karimanzira,\%20torsten.pfuetzenreuter\%7d@iosb-ast.fraunhofer.de}{@iosb-ast.fraunhofer.de}}} 
\maketitle
\begin{abstract}
Nowadays an accurate modeling of the system to be controlled is essential for reliable autopilot.\\
This paper presents a non-linear model of the autonomous underwater vehicle ``CWolf''. Matrices and the corresponding coefficients generate a parameterized representation for added mass, Coriolis and centripetal forces, damping, gravity and buoyancy, using the equations of motion, for all six degrees of freedom. The determination of actuator behaviour by surge tests allows the conversion of propeller revolutions to the respective forces and moments. Based on geometric approximations, the coefficients of the model can be specified by optimization algorithms in ``open loop'' sea trials.\\
The realistic model is the basis for the subsequent design of the autopilot. The reference variables used in the four decoupled adaptive PID controllers for surge, heading, pitch and heave are provided a ``Line of Sight'' - guidance system. A constraint criteria optimization determines the required controller parameters. The verification by ``closed loop'' sea trials  ensures the results.\\
\end{abstract}

%
\IEEEpeerreviewmaketitle
	\section{Introduction}
Fish farming in aquacultures is a sensible alternative to industrial fishing, which is accompanied by the loss of biodiversity, the exploitation of global fisheries and the increasing pollution of the oceans. In 2010 more than 60 million tons of fish, mussels and crabs came from such farms and about 600 species of animals are kept in aquacultures worldwide. The food sector, which has been growing fastest in recent years, has also it´s drawbacks. Feeding the animals and their excreta causes over-fertilization of the surrounding waters. Thus, the nutrient content increases in rivers, lakes and bays, destabilizing the natural ecosystem and endangering the local fauna and flora \cite{ExzellenzclusterOzeanderZukunft.2013}.
\begin{figure}[!b]
\vspace{-10pt}
\hrulefill\\
\footnotesize
\noindent\hspace*{10pt}This work was supported by the European Regional Development Fund (ERDF) of the European Union via the Thuringian Coordination Office TNA \#TNA VIII-1/2011
\vspace{-10pt}
\end{figure}
	\subsection{Project ``SALMON''}
Within the framework of the research project `` SALMON'' (Sea Water Quality Monitoring and Management), funded by the European Regional Development Fund (ERDF) \cite{FreistaatThuringen.2013}, an autonomous underwater vehicle (AUV) records and analyses water quality data. During planned missions in the vicinity of a fish farm, water samples are collected and analysed. In addition to the nitrate pollution, density, conductivity and salt concentration are determined \cite{Eichhorn.2013}.\\
The autopilot, the simulators and the mission planning software are developed by the Department of Systems Analysis at the Ilmenau University of Technology \cite{Ament.2013}. The construction and management of the AUV hardware components is conducted by the engineers at the Fraunhofer IOSB-AST \cite{Pfuetzenreuter.2013}. A sensor module records the measurement data, which is developed and maintained by 4H Jena GmbH \cite{Grunwald.2013}. The Norwegian Institute of Marine Research \cite{Toft.2013} provides the test environment around an experimental fish farm off the island of Austevoll in Hordaland (Norway) to carry out the final sea trials.
\begin{figure*}[!b]
	\centering
	\vspace*{5pt}
	\includegraphics[width=0.90\textwidth]{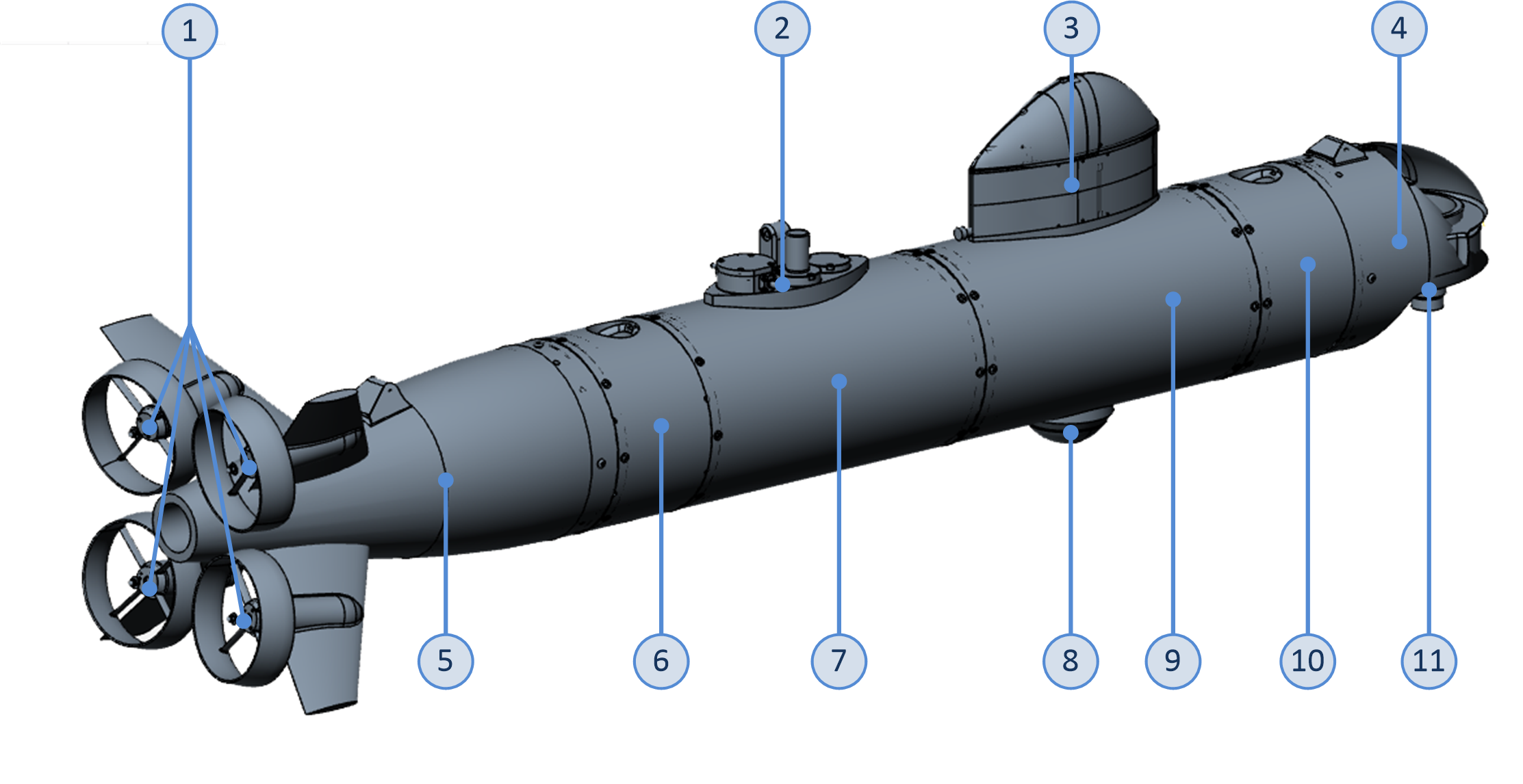}
	\caption{AUV ``CWolf'' outer layout}
	\label{221}
	\vspace*{-5pt}
\end{figure*}
	\subsection{AUV Base Platform ``CWolf''}
The autonomous underwater vehicle (AUV) ``CWolf'', whose performance parameters are shown in Table \ref{T01}, is based on the remote operated vehicle (ROV) ``SeaWolf'' from ATLAS Hydrographic GmbH \cite{Konnecke.2013}.
\begin{table}[!b]
		\vspace{-10pt}
		\renewcommand{\arraystretch}{1.3}
		\caption{performance parameter of AUV ``CWolf''}
		\label{T01}
		\centering
		\begin{tabular}{l r}
				\toprule	
				\bfseries Parameter 										& \bfseries Value					\\
				\midrule
				Length 																	& \ensuremath{2.50}~m			\\
				Diameter 																& \ensuremath{0.30}~m			\\
				Weight in air														& \ensuremath{135.00}~kg	\\
				Max. speed															& \ensuremath{6.00}~kn		\\
				Endurance at \ensuremath{3.00}~kn				& \ensuremath{3.00}~h			\\
				Payload																	& \ensuremath{15.00}~kg		\\
				\bottomrule
		\end{tabular}
\end{table}
After modifying it to an autonomous platform, the control system \textit{ConSys} is integrated \cite{Pfuetzenreuter.2010}. By means of the modular payload concept, different sensor systems can be installed inside and outside. Communication takes place with WLAN or fibre optic cable. Fibre optic gyroscope (FOG), Doppler Velocity Log (DVL), scanning sonar and Global Position System (GPS) detect the current position and orientation of the vehicle for navigation. The AUV is actuated by four stern propellers and two vertical thrusters. Different activations of individual propellers by steering and pitching commands allow the manoeuvring in horizontal and vertical plane. Thus, it is possible to pilot the vehicle without rudder units during forward movement and hovering. The vertical thrusters are used exclusively for vertical movement. \\
Figure \ref{221} shows the construction model of the AUV "CWolf" with components:
\begin{enumerate}[\IEEEsetlabelwidth{11.}]
		\item stern propellers
		\item sensor module
		\item Global Position System (GPS)
		\item bow section
		\item stern section
		\item vertical thruster stern
		\item payload section
		\item Doppler Velocity Log (DVL)
		\item vehicle guidance system (VGS)
		\item vertical thruster bow
		\item scanning sonar
\end{enumerate}
	\section{Modeling}
\begin{figure}[!h]
	\centering
	\vspace{-5pt}
	\includegraphics[width=0.48\textwidth]{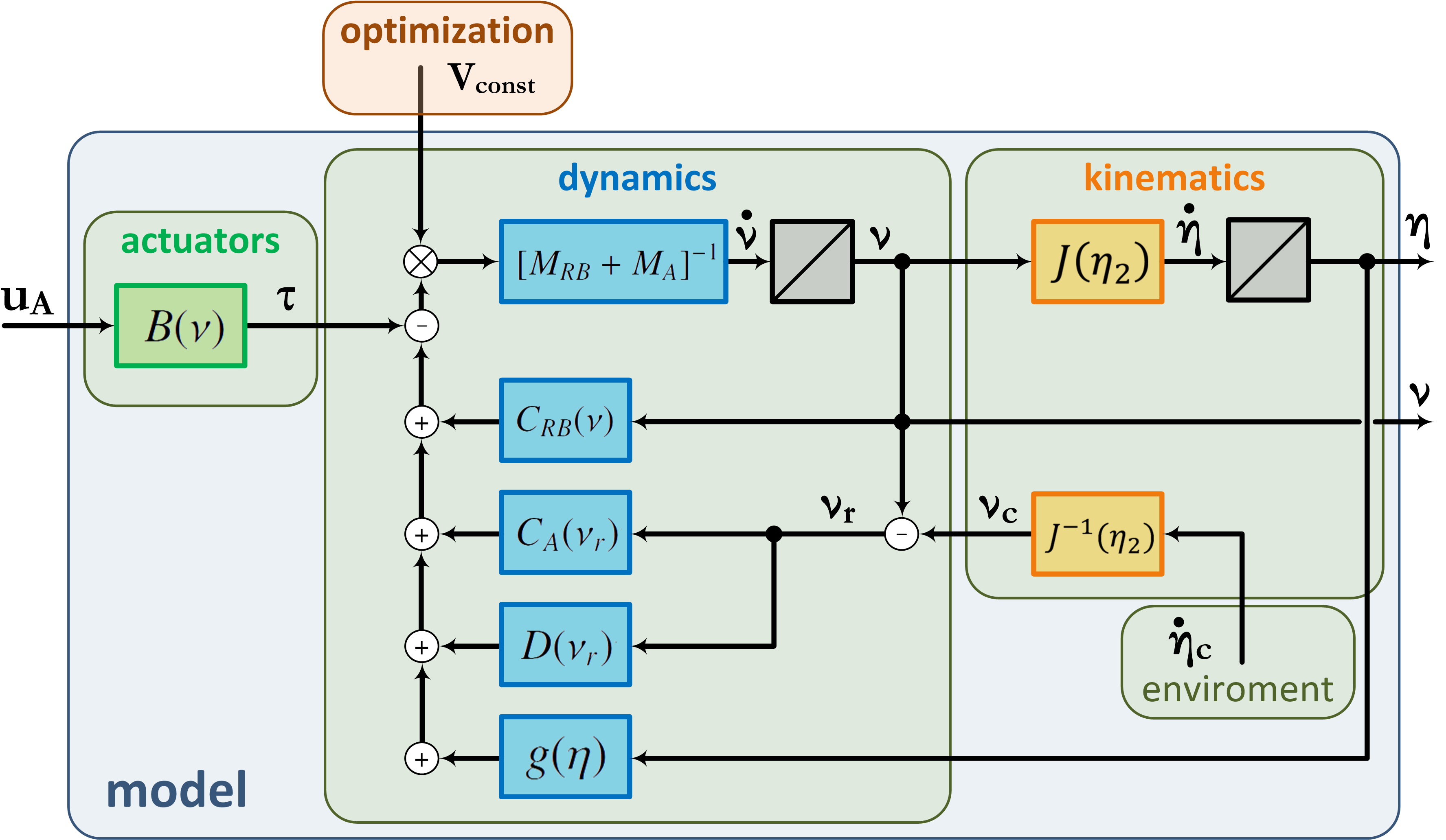}
	\caption{Block diagram of AUV model}
	\label{402}
	\vspace{-5pt}
\end{figure}

The non-linear continuous equation of motion (\ref{E210}) with six degrees of freedom (6DOF) serves as a mathematical description of the movement of an underwater vehicle in a fluid. The force-moment vector \ensuremath{\tau} summarizes the various physical influences as a function of acceleration \ensuremath{\dot{\nu}}, velocity \ensuremath{\nu} and position \ensuremath{\eta} \cite{Fossen.1994}.
\begin{IEEEeqnarray}{rCl}
\tau	&	=	& \left[X \quad Y	\quad Z \quad K 		\quad M 			\quad N 		\right]^{T} 	\\
\nu		&	=	&	\left[u \quad v	\quad w \quad p 		\quad q				\quad r 		\right]^{T}		\\
\eta 	&	=	&	\left[x \quad y	\quad z \quad \phi 	\quad \theta 	\quad \psi 	\right]^{T}		
\end{IEEEeqnarray}
The relative velocity of the vehicle \ensuremath{\nu_{r}} is used for calculation of the hydrodynamic terms. It's calculated by the subtracting the current velocity of the liquid \ensuremath{\nu_{c}} from the absolute velocity \ensuremath{\nu} \cite{Fossen.2006}.
\begin{IEEEeqnarray}{rCl}
\nu_{r}	&	=	& \nu - \nu_{c}	\\
				&	=	&	\left[u-u_{c} \quad v-v_{c}	\quad w-w_{c} \quad p \quad q \quad r \right]^{T}
\end{IEEEeqnarray}
Figure \ref{402} presents the schematic representation of the system model according to (\ref{E210}), which consists of several modules, explained in the following sections.
\begin{figure*}[!t]
\normalsize
\setcounter{MYtempeqncnt}{\value{equation}}
\setcounter{equation}{5}
\begin{equation}
	\label{E210}
	\dot{\nu} = [M_{RB} + M_{A}]^{-1} \cdot [B(\nu)~u_{A} - C_{RB}(\nu)\nu  - C_{A}(\nu_r)\nu_r - D(\nu_r)\nu_r - g(\eta)] \cdot V_{const}
\end{equation}
\setcounter{equation}{\value{MYtempeqncnt}}
\hrulefill
\vspace*{4pt}
\end{figure*}
	\subsection{Kinematics}
Depending on the available sensors, the navigation system provides position, orientation and motion data, which are determined in different reference systems. The transformation matrix is necessary for the conversion of the body-fixed (Bf) \ensuremath{\nu} frame values into earth-fixed (Ef) \ensuremath{\eta} frame values \cite{Fossen.2002}.
\begin{equation}
	\setcounter{equation}{7}
	\label{E211}
	\dot{\eta} = J(\eta)\nu
\end{equation}
	\subsection{Dynamics}
The hydrodynamic behaviour of the vehicle is described by matrices, which in combination with the movement vectors represent forces and moments. Simplifications, such as neglecting the coupling terms and the reduction to diagonal matrices, depend on various assumptions (eg. rotation-symmetric body shapes, low speed, high depth and insignificant current, etc.) \cite{Antonelli.2003}.
	\subsection*{Rigid-body inertia matrix \ensuremath{M_{RB}}}
The distribution of mass \ensuremath{m} in a rigid body and its relative distance to the center of gravity cause inertia and moments of inertia, which are combined in the constant matrix \ensuremath{M_{RB}}.
\begin{equation}
		M_{RB} = \left[\begin{IEEEeqnarraybox*}[][c]{,c/c,}
												m~I_{3\times3} & -m~S(r_{g})	\\
												m~S(r_{g})		 &	 I_{O}			%
									\end{IEEEeqnarraybox*}\right]
\end{equation}
The position vector \ensuremath{r_{g}} indicates the distance between the body-fixed origin (O) and the center of gravity (CG) of the vehicle. The calculation of the cross product of the mass and the position vector is computed by the skew-symmetric matrix \ensuremath{S(.)}. The 	inertia tensor \ensuremath{I_{O}} describes the moments of inertia with respect to all axes of the body.
\begin{equation}
		I_{O} = \left[\begin{IEEEeqnarraybox*}[][c]{,c/c/c,}
												 I_{x}		& -I_{xy}		& -I_{xz}	  \\
												-I_{yx}		&  I_{y}		& -I_{yz}  	\\
												-I_{zx}		& -I_{zy}		&  I_{z}  	%
									\end{IEEEeqnarraybox*}\right]
\end{equation}
Assuming that the center of gravity lies in the body-fixed origin (CG = O) and the deviation moments for rotation-symmetrical bodies are zero (\ensuremath{I_{xy}=I_{xz}=I_{yz}=0}), the inertia matrix is reduced to a diagonal matrix .
\begin{equation}
	M_{RB} = -diag\left\{m \quad m \quad m \quad I_{x} \quad I_{y} \quad I_{z}\right\}
\end{equation}
The geometrical calculation of a homogeneous straight cylinder approximates the outer shape of the vehicle. By extending the formula with the coefficients \ensuremath{C_{I_{x}}}, \ensuremath{C_{I_{y}}}, the subsequent optimization is adjusting the model.
\begin{IEEEeqnarray}{rCl}
		I_{x}	&	=	& \int_{V}(y^{2}+z^{2}) \cdot \rho_{M}~dV																				\IEEEnonumber \\
					&	= & C_{I_{x}} \cdot \rfrac{1}{2} \cdot m \cdot r^{2} 															\label{Ix}\\
		I_{y}	&	=	&	\int_{V}(x^{2}+z^{2}) \cdot \rho_{M}~dV 																			\IEEEnonumber \\
					&	=	& C_{I_{y}} \cdot \rfrac{1}{12}	\cdot m \cdot (l^{2} + 3 \cdot r^{2}) 					\label{Iy}\\
		I_{z}	&	=	& I_{y}																																					\label{Iz}
\end{IEEEeqnarray}
The geometric parameters are determined by the radius \ensuremath{r}, the length \ensuremath{l} and the volume \ensuremath{V} of the vehicle. For the material density \ensuremath{\rho_{M}}, a homogeneous distribution is assumed.
	\subsection*{Added mass matrix \ensuremath{M_{A}}}
By accelerating the vehicle, the surrounding fluid produces additional forces and moments. For underwater vehicles, a simplification  to diagonal matrix is done analogous to the inertia matrix.
\begin{equation}
	M_{A} = -diag\left\{X_{\dot{u}},Y_{\dot{v}},Z_{\dot{w}},K_{\dot{p}},M_{\dot{q}},N_{\dot{r}}\right\}
\end{equation}
By using a prolate ellipsoid to approach the form of the outer vehicle hull \cite{Fossen.1994}, a direct correlation between the added mass coefficients and vehicle mass (inertia) can be found. The factors \ensuremath{C_{X_{\dot{u}}}}, \ensuremath{C_{Y_{\dot{v}}}} \ensuremath{C_{Z_{\dot{w}}}}, \ensuremath{C_{M_{\dot{q}}}} and \ensuremath{C_{N_{\dot{r}}}} are optimized.
\begin{IEEEeqnarray}{rCl}
		X_{\dot{u}}	&	=	& -C_{X_{\dot{u}}} (\rfrac{2}{3} \pi \rho_{F} r^{2} l) = -C_{X_{\dot{u}}} m \\
		Y_{\dot{v}}	&	= & -C_{Y_{\dot{v}}} (\rfrac{2}{3} \pi \rho_{F} r^{2} l) = -C_{Y_{\dot{v}}} m	\\
		Z_{\dot{w}}	&	=	&	-C_{Z_{\dot{w}}} (\rfrac{2}{3} \pi \rho_{F} r^{2} l) = -C_{Z_{\dot{w}}} m	\\[5pt]		
		M_{\dot{q}}	&	=	& -C_{M_{\dot{q}}} (\rfrac{2}{15} \pi \rho_{F} r^{2} l) (\rfrac{1}{4} l^{2} + r^{2})	= -C_{M_{\dot{q}}} I_{y}	\\
		N_{\dot{r}}	&	=	& -C_{N_{\dot{r}}} (\rfrac{2}{15} \pi \rho_{F} r^{2} l) (\rfrac{1}{4} l^{2} + r^{2})	= -C_{N_{\dot{r}}} I_{z}
\end{IEEEeqnarray}
The added mass for roll cannot be used for the approach of prolate ellipsoid (\ensuremath{K_{\dot{p}} \neq 0}). Due to additional structures at the outer hull of the vehicle (fins and tower), a factor \ensuremath{C_{K_{\dot{p}}}} must be introduced.
\begin{equation}
		K_{\dot{p}}	=	-C_{K_{\dot{p}}} m
\end{equation}
	\subsection*{Rigid-body Coriolis and centripetal matrix \ensuremath{C_{RB}(\nu)}}
The rigid body Coriolis and centripetal terms are derived from the rigid body inertia matrix \ensuremath{M_{RB}}. They are connected to the body-fixed velocity of the vehicle \ensuremath{\nu} as in \cite{Fossen.2002}.
\begin{equation}
		C_{RB}(\nu) = \left[\begin{IEEEeqnarraybox*}[\mysmallarraydecl][c]{,c/c,}
															0_{3\times3} 														& -m S(\nu_{1}) - m S(\nu_{2}) S(r_{g})	\\
															-m S(\nu_{1}) + m S(r_{g}) S(\nu_{2}) 	&	-S(I_{O}\nu_{2})											%
													\end{IEEEeqnarraybox*}\right]
\end{equation}
For a concentrated representation the body-fixed velocity vector \ensuremath{\nu} is divided in \ensuremath{\nu_{1} = \left[u ~~ v	\quad w \right]^{T}} and  \ensuremath{\nu_{2} = \left[ p \quad q				\quad r\right]^{T}}.
	\subsection*{Hydrodynamic Coriolis and centripetal matrix \ensuremath{C_{A}(\nu_{r})}}
The hydrodynamic Coriolis and centripetal terms also result from the added mass matrix \ensuremath{M_{A}} and are linked with the relative body fixed velocity \ensuremath{\nu_{r}} via cross product.
\begin{equation}
		C_{A}(\nu_{r}) = \left[\begin{IEEEeqnarraybox*}[\mysmallarraydecl][c]{,c/c/c/c/c/c,}
															0									&  0								&	 0								&	 0								&	-Z_{\dot{w}}w_{r}	&	 Y_{\dot{v}}v_{r}	\\
															0									&  0								&	 0								&	 Z_{\dot{w}}w_{r}	&	 0								&	-X_{\dot{u}}u_{r}	\\
															0									&  0								&	 0								&	-Y_{\dot{v}}v_{r}	&	 X_{\dot{u}}u_{r}	&	 0								\\
															0									& -Z_{\dot{w}}w_{r}	&	 Y_{\dot{v}}v_{r}	&	 0								&	-N_{\dot{r}}r			&	 M_{\dot{q}}q			\\
															Z_{\dot{w}}w_{r}	&  0								&	-X_{\dot{u}}u_{r}	&	 N_{\dot{r}}r			&	 0								&	-K_{\dot{p}}p			\\
															-Y_{\dot{v}}v_{r}	&  X_{\dot{u}}u_{r}	&  0								& -M_{\dot{q}}q			&  K_{\dot{p}}p			&  0
													\end{IEEEeqnarraybox*}\right]
\end{equation}
	\subsection*{Damping matrix \ensuremath{D(\nu_{r})}}
The hydrodynamic damping, that acts on the vehicle, can be reduced to linear and quadratic terms. The influence of wind and waves is neglected. Only the energy loss due to friction (laminar and turbulent) and the viscosity of the water should be considered at this point.
\begin{IEEEeqnarray}{rCl}
	D(\nu_{r}) & = & - diag\left\{X_{u},Y_{v},Z_{w},K_{p},M_{q},N_{r}\right\} 														\\
						 &	 & - diag\left\{X_{u|u|},Y_{v|v|},Z_{w|w|},K_{p|p|},M_{q|q|},N_{r|r|}\right\}|\nu_{r}| 	\IEEEnonumber
\end{IEEEeqnarray}
The linear terms are neglected in most publications (eg. \cite{Ferreira.2009}), because they are minuscule in the forward velocity with range of \ensuremath{0,5 \leq u \leq 2,5~\rfrac{m}{s}}. The following optimization of these parameters shows that this assumption is not useful in all cases. \\
In general the square damping term depends on the material density \ensuremath{\rho_{F}}, the drag coefficient \ensuremath{C_{D}} and the projection area. The quadratic damping coefficient by frontal flow \ensuremath{X_{u|u|}} is determined by the surface in yz-plane \ensuremath{A_{yz}}.
\begin{equation}
		X_{u|u|} = - \rfrac{1}{2} \cdot \rho_{F} \cdot C_{D_{x}} \cdot A_{yz}
\end{equation}
Assuming an ellipsoidal hull shape (projection area: \ensuremath{A_{yz} = \pi \cdot r^{2}}) and applying the laminar flow theory, the drag coefficient \ensuremath{C_{D_{x}}} can be identified by the friction coefficient \ensuremath{C_{f}} \cite{Hoerner.1965}.
\begin{equation}
		C_{D_{x}} = 0,44 \frac{2r}{l} + 4 C_{f} \frac{l}{2r} + 4 C_{f}  \left(\frac{2r}{l}\right)^{\rfrac{1}{2}}
\end{equation}
With knowledge of the Reynolds number \ensuremath{Re}, the friction coefficient \ensuremath{C_{f}} is extracted from the ITTC-curve \cite{Lewis.1988}. 
\begin{equation}
		C_{f} = \frac{0,075}{(\log_{10}Re-2)^{2}}
\end{equation}
The damping in the x- and y-direction is approximated by the lateral flow of a short circular cylinder (\ensuremath{A_{xy} = A_{xz} = 2 \cdot r \cdot l}) \cite{White.op.2011}.
\begin{IEEEeqnarray}{rCl}
		Y_{v|v|} & = & - \rfrac{1}{2} \cdot \rho_{F} \cdot C_{D_{y}} \int_{-\rfrac{l}{2}}^{\rfrac{l}{2}} 2r~dx \IEEEnonumber\\
						 & = & - \rfrac{1}{2} \cdot \rho_{F} \cdot C_{D_{y}} \cdot A_{xz} 																					\\
		Z_{w|w|} & = & - \rfrac{1}{2} \cdot \rho_{F} \cdot C_{D_{z}} \int_{-\rfrac{l}{2}}^{\rfrac{l}{2}} 2r~dx \IEEEnonumber\\
						 & = & - \rfrac{1}{2} \cdot \rho_{F} \cdot C_{D_{z}} \cdot A_{xy}	%
\end{IEEEeqnarray}
To calculate the damping moments a homogeneous straight circular cylinder is used as basis. For this application, an additional drag coefficient for the rotation must be introduced \cite{Prestero.27.11.2001}.
\begin{IEEEeqnarray}{rCl}
		M_{q|q|} & = & - \rfrac{1}{2} \cdot \rho_{F} \cdot C_{D_{z}} \cdot C_{D_{q}} \cdot \int_{-\rfrac{l}{2}}^{\rfrac{l}{2}}|x|^{3} \cdot 2r~dx 	\IEEEnonumber \\
						 & = & - \rfrac{1}{32} \cdot \rho_{F} \cdot C_{D_{z}} \cdot C_{D_{q}} \cdot r \cdot l^{~4}																												\\
		N_{r|r|} & = & - \rfrac{1}{2} \cdot \rho_{F} \cdot C_{D_{y}} \cdot C_{D_{r}} \cdot \int_{-\rfrac{l}{2}}^{\rfrac{l}{2}}|x|^{3} \cdot 2r~dx 	\IEEEnonumber \\
						 & = & - \rfrac{1}{32} \cdot \rho_{F} \cdot C_{D_{y}} \cdot C_{D_{r}} \cdot r \cdot l^{~4}
\end{IEEEeqnarray}
Analogous to the calculations of the added mass the rotation about the x-axis cannot be neglected (\ensuremath{K_{p|p|} \neq 0}). Therefore, the projection areas of additional structures on the outer hull are summarized in \ensuremath{A_{R} = 4	\cdot \underbrace{(2r_{B} \cdot r_{B})}_{\text{fin}} + \underbrace{2 \cdot r_{B} \cdot r_{B}}_{\text{tower}}}.
\begin{equation}
		K_{p|p|} = - \rfrac{1}{2} \cdot \rho_{F} \cdot C_{D_{p}} \cdot A_{R}
\end{equation}	
	\subsection*{Gravity and buoyancy matrix \ensuremath{g(\eta)}}
The weight \ensuremath{W} and the buoyancy \ensuremath{B} of all vehicle sections are critical for calculating the restoring forces and moments \ensuremath{g(\eta)}. The transformation \ensuremath{J_1^{-1}(\eta_2)} converts the body-fixed values to the required earth-fixed relationships.
\begin{IEEEeqnarray}{rCl}
	g(\eta)	& = &	- \left[\begin{IEEEeqnarraybox*}[\mysmallarraydecl][c]{,c,}
															J_1^{-1}(\eta_2)~\left(\begin{IEEEeqnarraybox*}[\mysmallarraydecl][c]{,c,}
																														0	\\
																														0 \\
																														W	
																											\end{IEEEeqnarraybox*}\right) + J_1^{-1}(\eta_2)~\left(\begin{IEEEeqnarraybox*}[\mysmallarraydecl][c]{,c,}
																																																												0	\\
																																																												0 \\
																																																												B	
																																																							\end{IEEEeqnarraybox*}\right)												\\[3pt]
															r_g \times J_1^{-1}(\eta_2)~\left(\begin{IEEEeqnarraybox*}[\mysmallarraydecl][c]{,c,}
																																						0	\\
																																						0 \\
																																						W	
																																\end{IEEEeqnarraybox*}\right) + r_b \times J_1^{-1}(\eta_2)~\left(\begin{IEEEeqnarraybox*}[\mysmallarraydecl][c]{,c,}
																																																																					0	\\
																																																																					0 \\
																																																																					B	
																																																																	\end{IEEEeqnarraybox*}\right)
												\end{IEEEeqnarraybox*}\right]
\end{IEEEeqnarray}
Since the respective forces apply at the center of gravity \ensuremath{r_g = \left[x_g ~~ y_g ~~ z_g\right]^{T}} and the center of buoyancy \ensuremath{r_b = \left[x_b ~~ y_b ~~ z_b\right]^{T}}, the exact location of the centers must be known.
\begin{IEEEeqnarray}{rCl}
		g(\eta)	& = & - \left[\begin{IEEEeqnarraybox*}[\mysmallarraydecl][c]{,c,}
													 (W-B) \sin{\theta}																																\\
													-(W-B) \cos{\theta}\sin{\phi}																											\\
													-(W-B) \cos{\theta}\cos{\phi}																											\\
													-(y_{g}W-y_{b}B) \cos{\theta}\cos{\phi} + (z_{g}W-z_{b}B) \cos{\theta}\sin{\phi}	\\
													 (z_{g}W-z_{b}B) \sin{\theta} + (x_{g}W-x_{b}B) \cos{\theta}\cos{\phi}						\\
													-(x_{g}W-x_{b}B) \cos{\theta}\sin{\phi} + (y_{g}W-y_{b}B) \sin{\theta}						
													\end{IEEEeqnarraybox*}\right]
\end{IEEEeqnarray}
	\subsection{Actuators}
\begin{figure}[!h]
	\centering
	\vspace{-5pt}
	\includegraphics[width=0.48\textwidth]{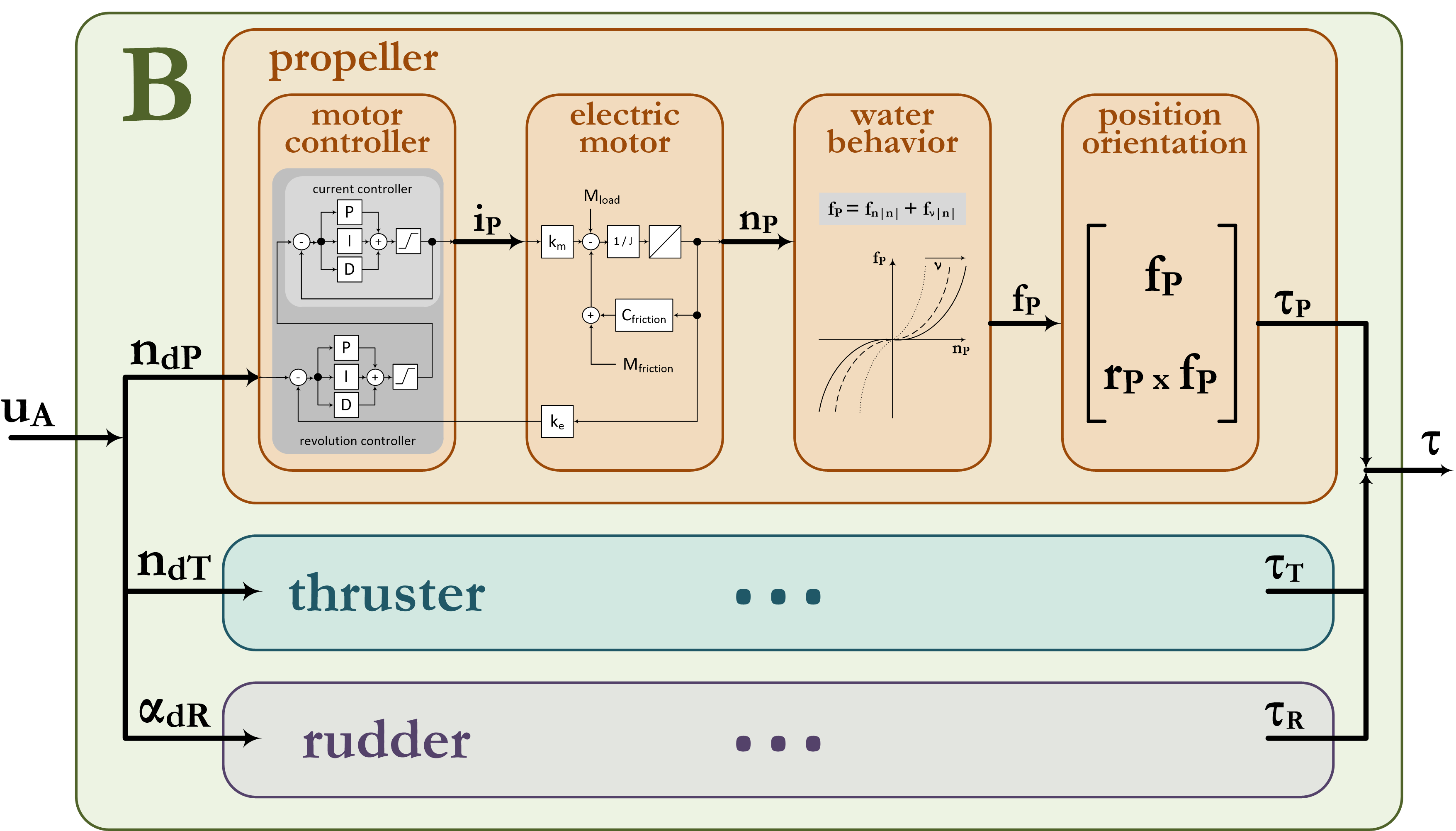}
	\caption{Block diagram of actuator matrix}
	\label{325}
	\vspace{-15pt}
\end{figure}

The central element of modeling with the 6DOF - equation of motion is the representation of all effects in the force-moment vector \ensuremath{\tau}. A control vector \ensuremath{u_{A}}, containing for example desired revolution of the propellers \ensuremath{n_{dP}}, serves as an interface to the actuators (propellers, thrusters, rudders, etc.) of the vehicle. \\
The Matrix \ensuremath{B} (Fig. \ref{325}) transforms the desired values into the generated forces and moments. For the transformation the static/dynamic behaviour of the motor controller, the characteristic curve of the propeller in water and the position/orientation of the actuator need to be considered.
	\subsection*{Static behaviour}
The motor controller limits the desired revolution speed \ensuremath{n_{d}} to a negative \ensuremath{n_{max}^{-}} and positive maximum \ensuremath{n_{max}^{+}}. In the revolution speed range between a negative \ensuremath{n_{min}^{-}} and positive low limit \ensuremath{n_{min}^{+}} no stable values for actual revolution speed \ensuremath{n_{a}} ​​can be guaranteed.
\begin{figure}[!ht]
	\centering
	\vspace{-5pt}
		\begin{minipage}[c]{0.26\textwidth}
						\vspace{0pt}
						\centering
						\includegraphics[width=0.99\textwidth]{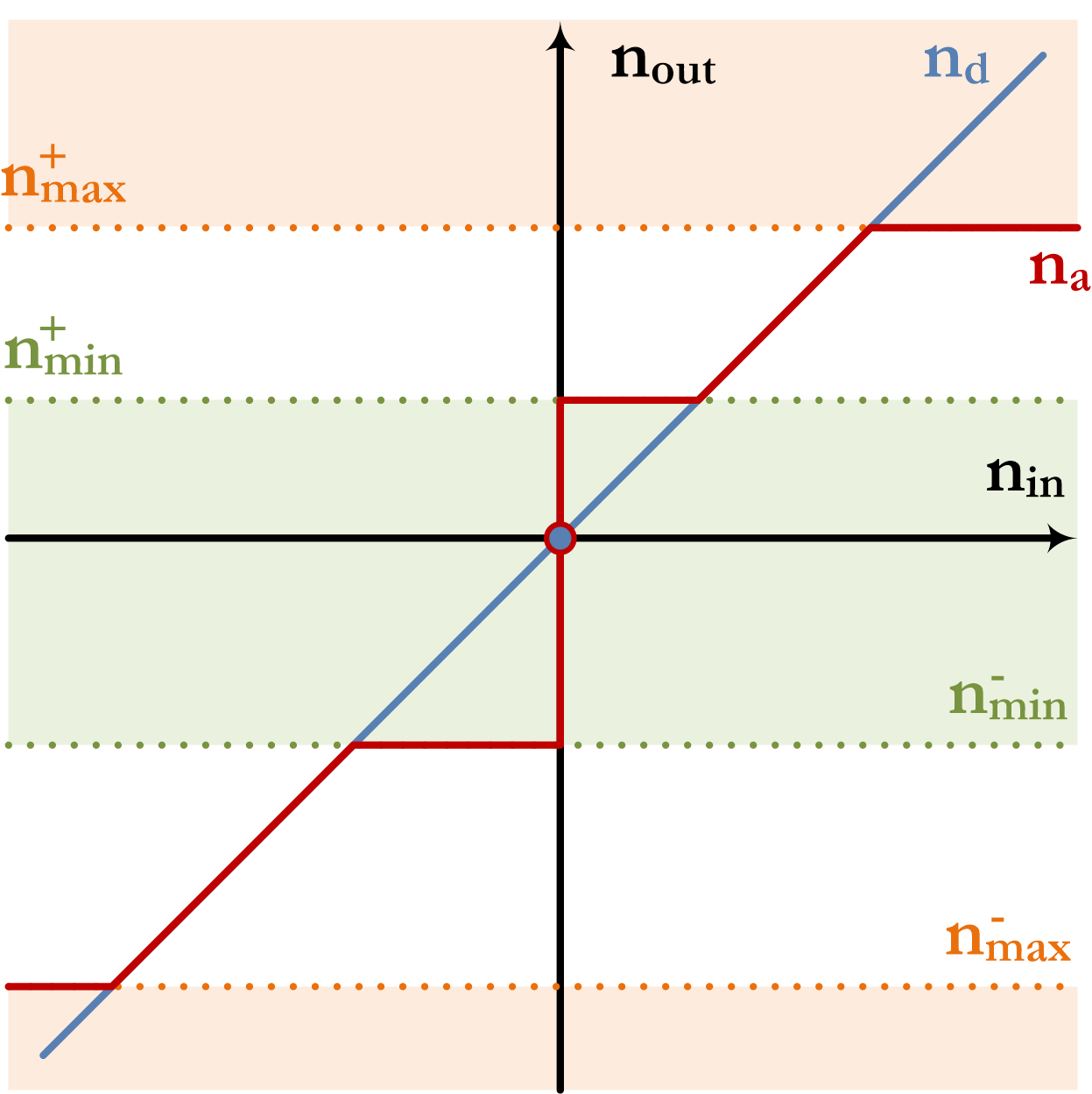}
						\vspace{-0pt}
		\end{minipage}
		\begin{minipage}[c]{0.22\textwidth}
						\vspace{0pt}
						\centering
						\begin{equation}
								\setlength{\nulldelimiterspace}{0pt}
								n_{a} = \left\{\begin{IEEEeqnarraybox}[\mysmallarraydecl][c]{l?s}
																n_{max}^{+},&for $n_{d} \geq n_{max}^{+}$					\\
																n_{min}^{+},&for $n_{d} > 0$											\\
																						&and $n_{d} < n_{min}^{+}$						\\
																0,					&for $n_{d} = 0$											\\
																n_{min}^{-},&for $n_{d} < 0$											\\
																						&and $n_{d} > n_{min}^{-}$						\\
																n_{max}^{-},&for $ n_{d} \leq n_{max}^{-}$				\\
																n_{d},			&otherwise
																\end{IEEEeqnarraybox}\right.			\IEEEnonumber
						\end{equation}
						\vspace{0pt}
		\end{minipage}
	\caption{Static transfer behaviour of the motor controller}
	\label{4631}
	\vspace{-5pt}
\end{figure}
	\subsection*{Dynamic behaviour}
The increase \ensuremath{n_{acc}} and decrease \ensuremath{n_{dec}} of the desired revolution speed is limited to reduce the load by high motor currents due to rapid and large revolution speed changes.\\
\begin{figure}[!h]
	\centering
	\vspace{-10pt}
	\includegraphics[width=0.45\textwidth]{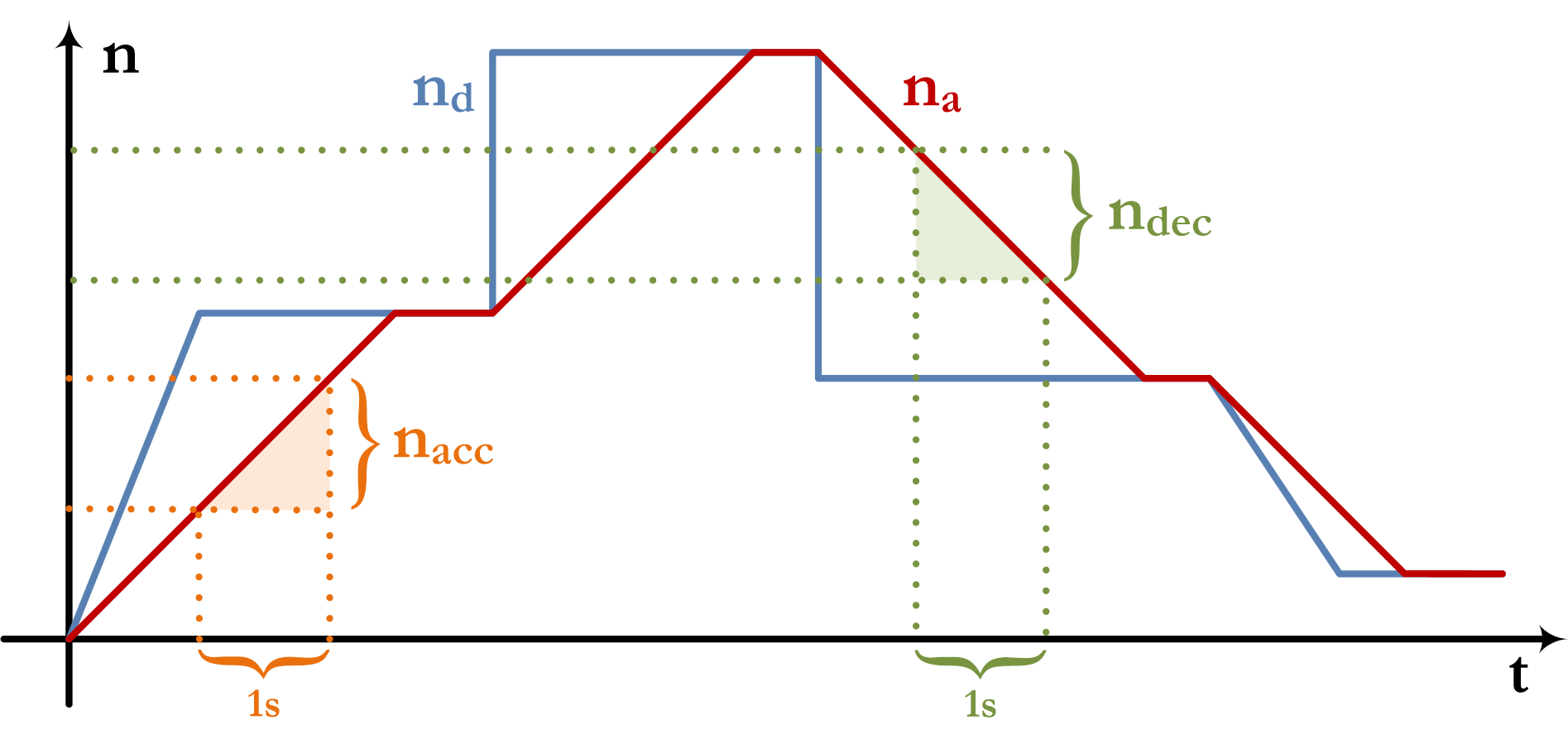}
	\caption{Dynamic limiting of increasing and decreasing}
	\label{4632}
	\vspace{-5pt}
\end{figure}
In addition, the dynamic transfer behaviour of the closed control loop of the motor controller system leads to a delay in the time domain. The command action of the PID controller can be approximated using a \ensuremath{PT_{1}} element with time delay.
\begin{equation}
		G(s)  = \frac{K_{V}}{(T_{1}s+1)}~e^{-T_{t}s}
\end{equation}	
\begin{figure}[!h]
	\centering
	\vspace{-5pt}
	\includegraphics[width=0.48\textwidth]{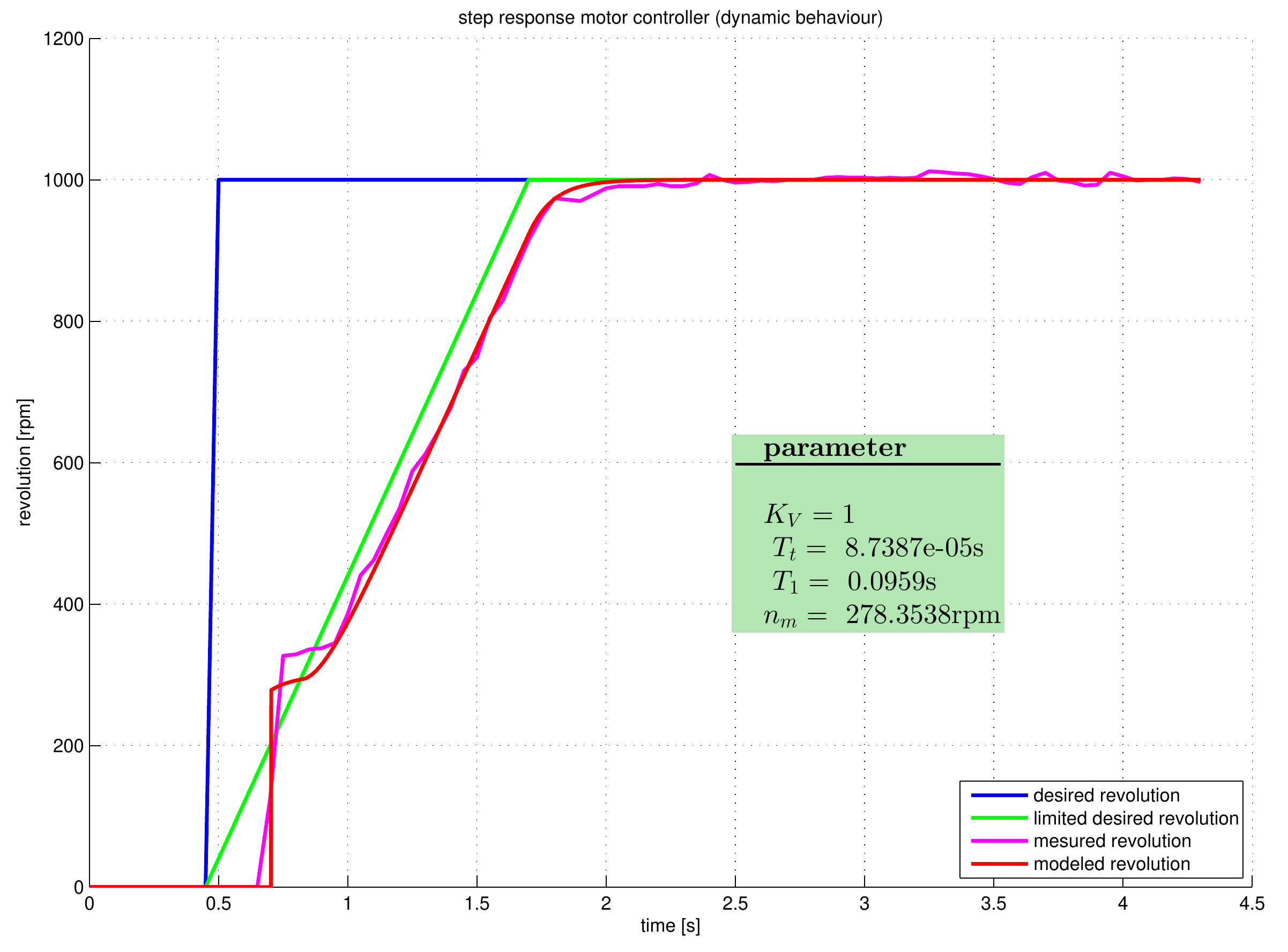}
	\caption{Modelled and measured step response of revolution}
	\label{4634}
	\vspace{-15pt}
\end{figure}
	\subsection*{Motor characteristic}
\begin{figure}[!h]
	\centering
	\vspace{-5pt}
	\includegraphics[width=0.48\textwidth]{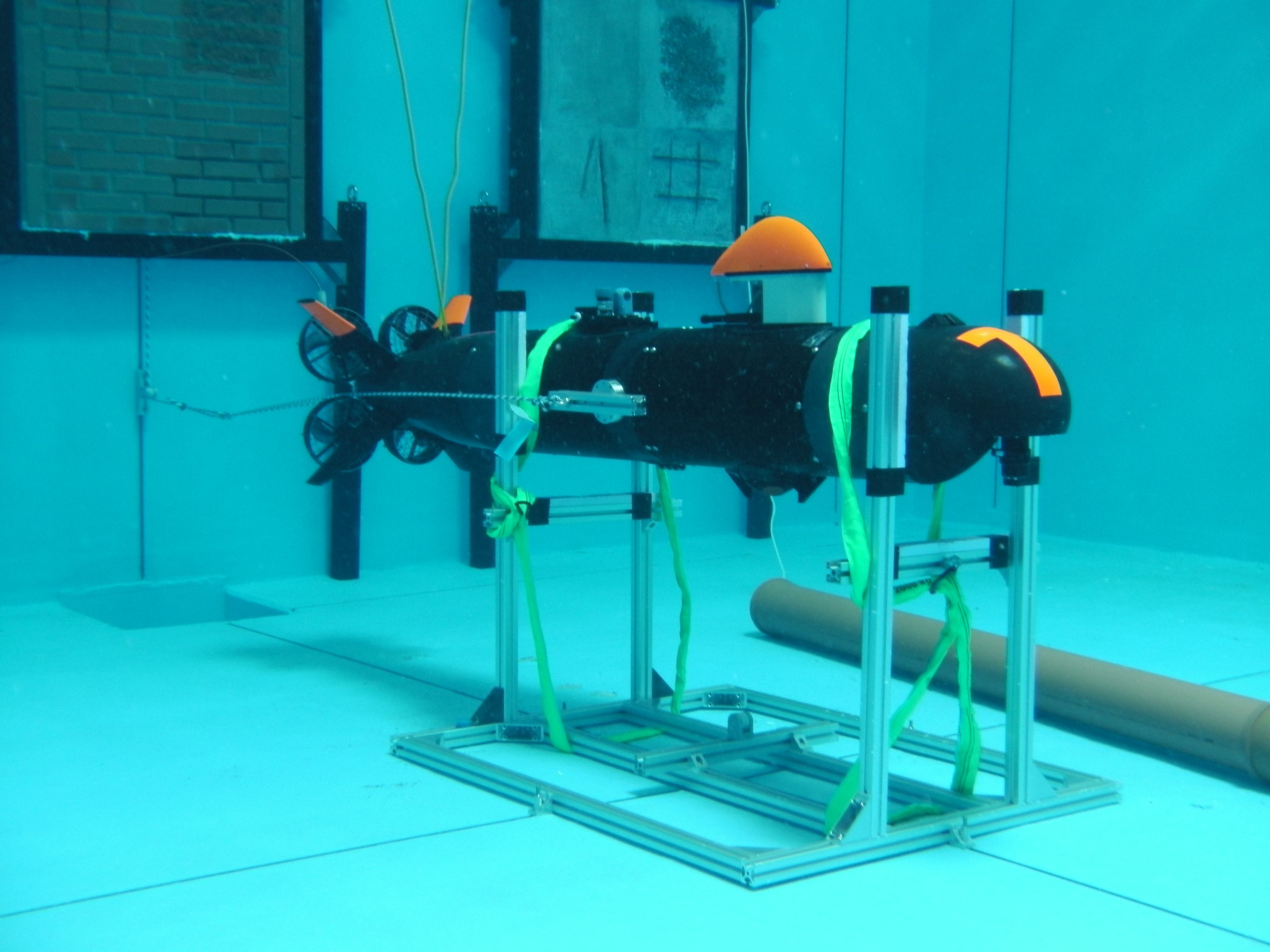}
	\caption{Surge test}
	\label{4644}
	\vspace{-5pt}
\end{figure}

\begin{figure}[!t]
	\centering
	\vspace{0pt}
	\includegraphics[width=0.48\textwidth]{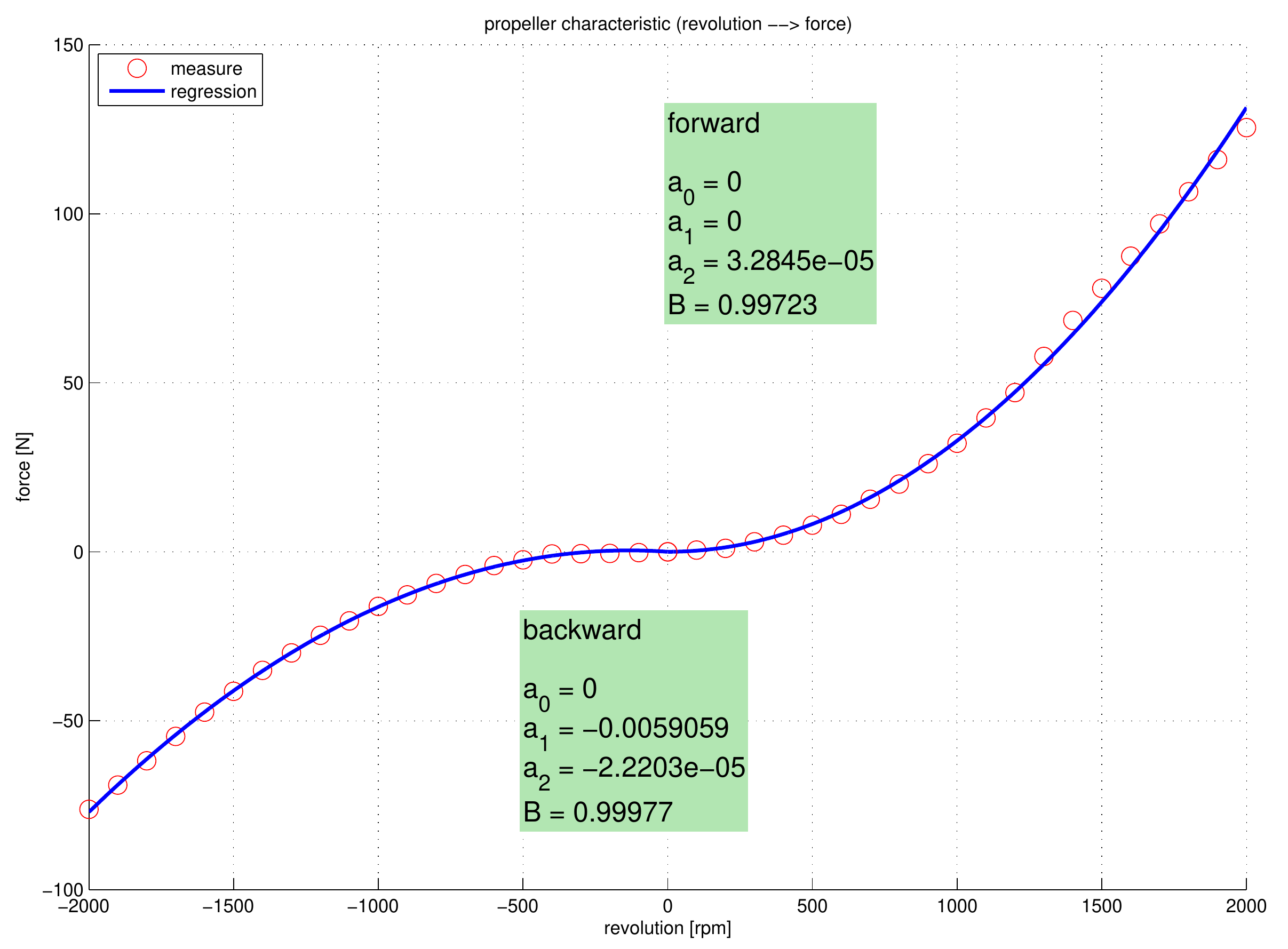}
	\caption{Motor characteristic of stern propeller estimated by direct regression}
	\label{4648}
	\vspace{-5pt}
\end{figure}

\begin{figure}[!t]
	\centering
	\vspace{0pt}
	\includegraphics[width=0.48\textwidth]{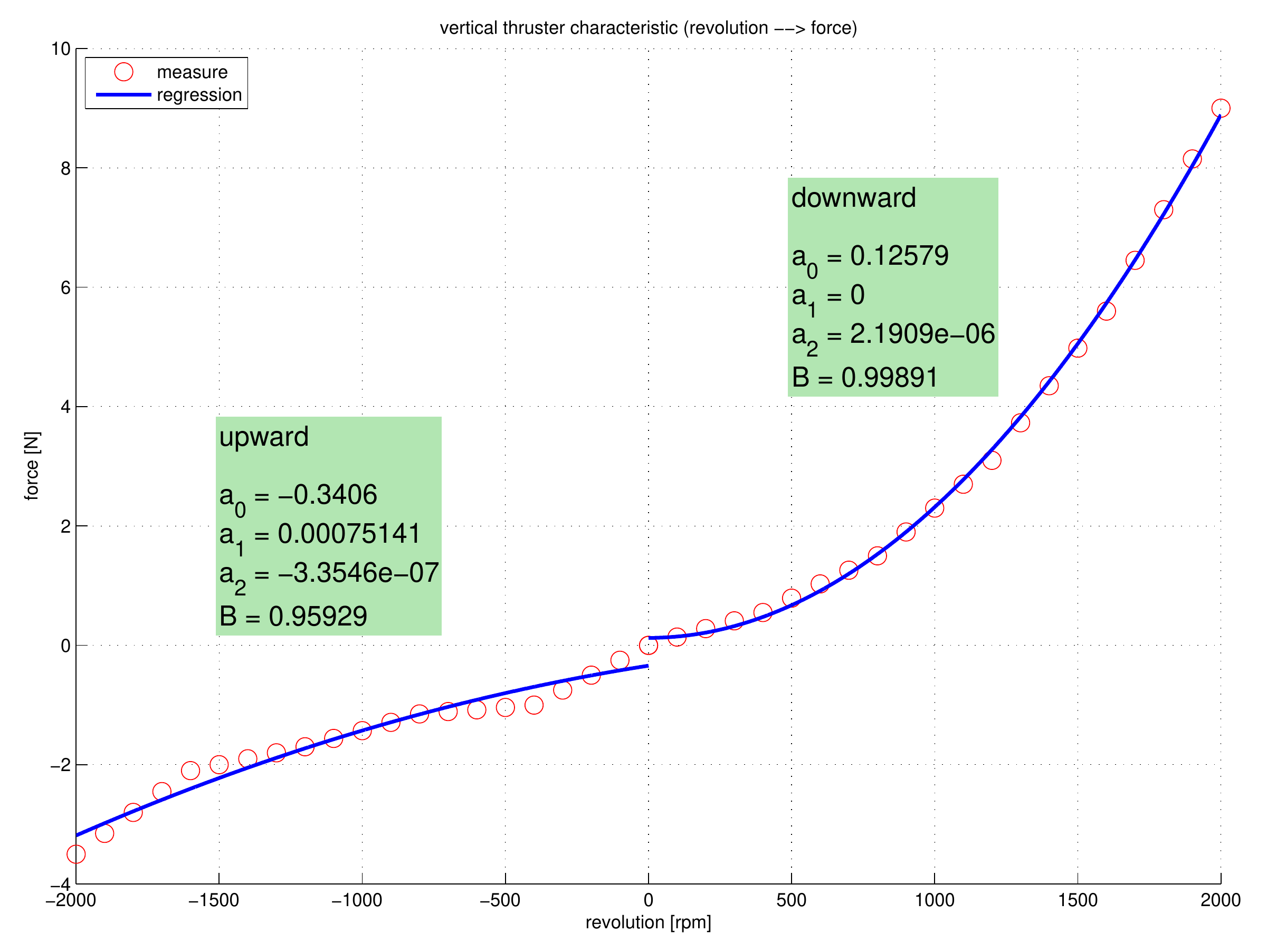}
	\caption{Motor characteristic of vertical thruster estimated by direct regression}
	\label{4649}
	\vspace{-5pt}
\end{figure}

The relationship between the actual revolution speed \ensuremath{n_{P}} and the propulsive force generated \ensuremath{F_{P}} is declared by the characteristic curve, which reflects the behaviour of the propeller in the water. According to \cite{Fossen.1994} this non-linear curve can be represented as a polynomial of second order. The coefficients \ensuremath{\alpha_{1}}, \ensuremath{\alpha_{2}} and the diameter of the propeller \ensuremath{d_{P}} describe the hydrodynamic and structural properties. The density \ensuremath{\rho_{F}} and the velocity of the fluid \ensuremath{V_{a}} are also included in the equation.
\begin{equation}
		F_{P} = \underbrace{\alpha_{1} \cdot \rho_{F} \cdot d_{P}~^{4}}_{p_{1}} \cdot |n_{P}|n_{P} + \underbrace{\alpha_{2} \cdot \rho_{F} \cdot d_{P}~^{3} \cdot V_{a}}_{p_{2}} \cdot |n_{P}|
\end{equation}	
This quadratic relationship is determined by surge tests. For this purpose, the AUV was fixed in a suitable depth of a test basin with sufficient distance from the edge of the pool, to prevent the vertical and horizontal pulling of the vehicle during acceleration (Fig. \ref{4644}). 
A load cell measures the over deflection pulleys and haulage ropes acting force generated by the respective actuators. In four experiments, the entire revolution range of all actuators in both directions can be scanned:
\begin{itemize}
	\item positive x-direction (stern propeller forward)
	\item negative x-direction (stern propeller backward)
	\item positive z-direction (vertical thruster downward)
	\item negative z-direction (vertical thruster upward)
\end{itemize}
Afterwards the total forces obtained must be separated into the individual components of each actuator (see section ``position and orientation''). The desired quadratic relationship between revolution (measured value matrix \ensuremath{U_{R}}) and force of a single actuator (output signals \ensuremath{y_{R}}) is determined by the direct regression using a second order model.
\begin{equation}
		y_{R} = a_{0} + \sum^{k}_{i=1}a_{i}u_{i} + \sum^{k-1}_{i=1}\sum^{k}_{j=i+1}a_{ij}u_{i}u_{j} + \sum^{k}_{i=1}a_{ii}u_{i}^{2}
\end{equation}	
\begin{equation}
		\hat{a}_{R} =  [U_{R}^{T}~U_{R}]^{-1}~U_{R}^{T}~y_{R}
\end{equation}	
Due to the negative and positive revolution speed behaviour, a parameter vector \ensuremath{\hat{a}_{R}} is estimated for the propellers and the vertical thrusters in both directions. The coefficient of determination \ensuremath{B} evaluates the quality of the regression from the 	correlation of the measured and calculated values​​.
\begin{equation}
		B = Kor(\hat{y},y)^{2} = \left(\frac{Cov(\hat{y},y)}{\sigma_{\hat{y}}~\sigma_{y}}\right)^{2} = \frac{\sum(\hat{y}-\bar{y})^{2}}{\sum(y-\bar{y})^{2}}
\end{equation}	
Based on the determined motor characteristics for the stern propellers (Fig. \ref{4648}) and the vertical thrusters (Fig. \ref{4649}),  it is possible to assign a force generated by the actuator to any actual revolution speed in the range \ensuremath{-2000...2000~}rpm 
	\subsection*{Position and orientation}
\begin{figure}[!h]
		\vspace{0pt}
		\centering
		\includegraphics[width=0.48\textwidth]{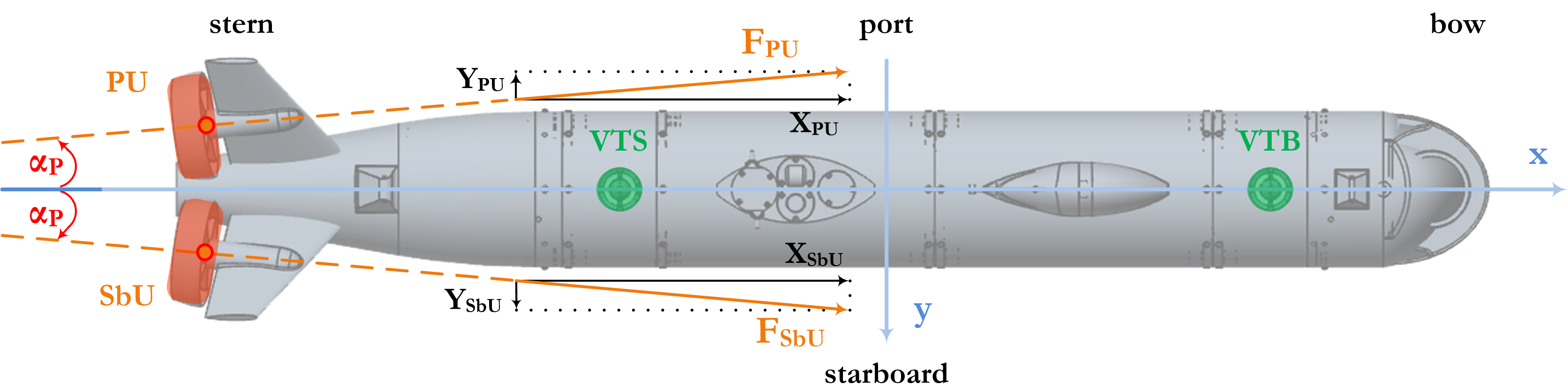}
		\vspace{10pt}
		\includegraphics[width=0.48\textwidth]{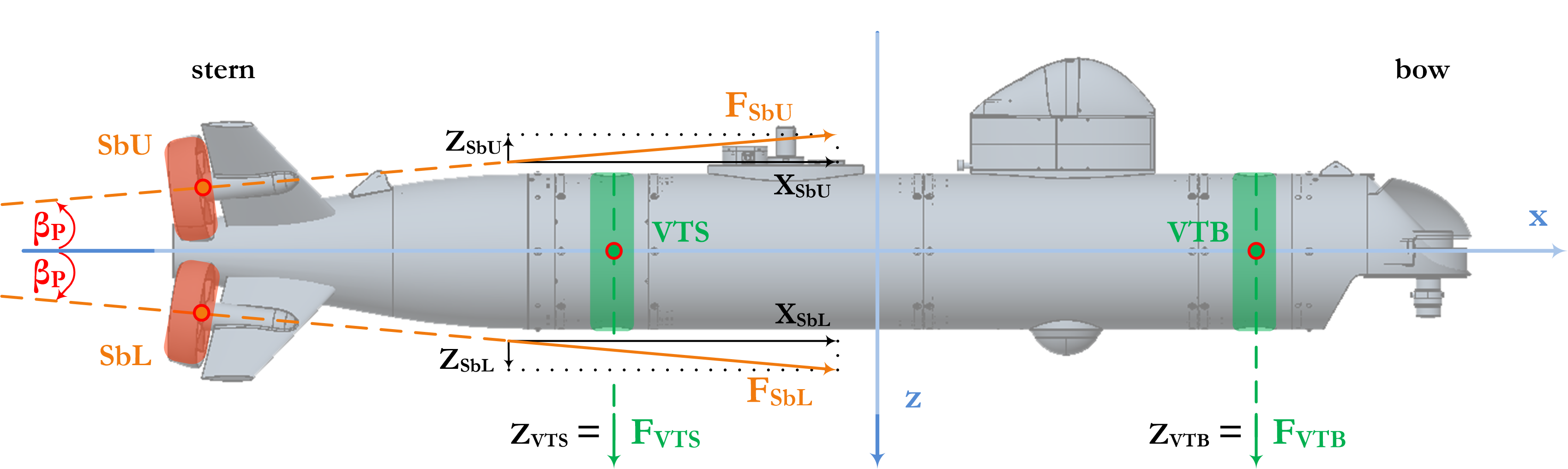}
		\vspace{10pt}
		\includegraphics[width=0.24\textwidth]{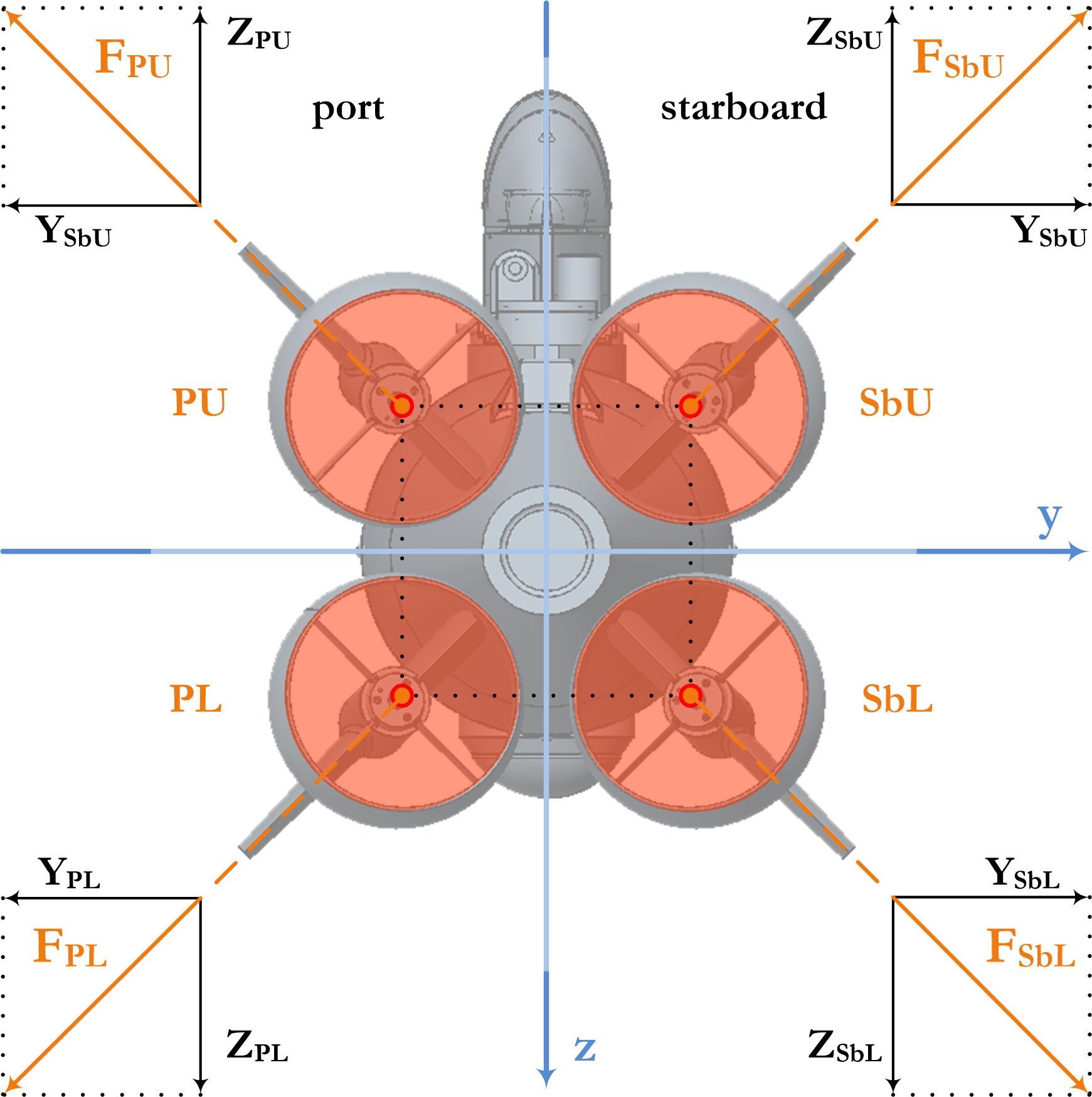}
	\caption{Force direction of stern propellers and vertical thrusters}
	\label{465}
	\vspace{-10pt}
\end{figure}

The force-moment vector \ensuremath{\tau} depends on the arrangement of the actuators with respect to the vehicle center of gravity.\\ 
The individual forces of the stern propellers (\ensuremath{F_{SbU}}, \ensuremath{F_{SbL}}, \ensuremath{F_{PU}} and \ensuremath{F_{PL}}) and of the vertical thrusters (\ensuremath{F_{VTB}} and \ensuremath{F_{VTS}}), determined by the surge test, act in the directions shown in Fig. \ref{465}. The inclination angle of the propellers \ensuremath{\alpha_{P}} and \ensuremath{\beta_{P}} complicate the trigonometric relationships.
\begin{IEEEeqnarray}{rCl}
		f_{SbU}	& = & F_{SbU} \cdot \left[\begin{IEEEeqnarraybox*}[\mysmallarraydecl][c]{,c,}
																								+~\sin{(90^{\circ}-\beta_{P})} \cdot \sin{(90^{\circ}-\alpha_{P})} \\	
																								+~\sin{(90^{\circ}-\beta_{P})} \cdot \cos{(90^{\circ}-\alpha_{P})} \\	
																								-~\cos{(90^{\circ}-\beta_{P})}
																			\end{IEEEeqnarraybox*}\right] 																								\\[3pt]
		f_{SbL}	& = & F_{SbL} \cdot \left[\begin{IEEEeqnarraybox*}[\mysmallarraydecl][c]{,c,}
																								+~\sin{(90^{\circ}-\beta_{P})} \cdot \sin{(90^{\circ}-\alpha_{P})} \\
																								+~\sin{(90^{\circ}-\beta_{P})} \cdot \cos{(90^{\circ}-\alpha_{P})} \\
																								+~\cos{(90^{\circ}-\beta_{P})}
																			\end{IEEEeqnarraybox*}\right] 																								\\[3pt]	
		f_{PU}	& = & F_{PU} \cdot  \left[\begin{IEEEeqnarraybox*}[\mysmallarraydecl][c]{,c,}
																								+~\sin{(90^{\circ}-\beta_{P})} \cdot \sin{(90^{\circ}-\alpha_{P})} \\
																								-~\sin{(90^{\circ}-\beta_{P})} \cdot \cos{(90^{\circ}-\alpha_{P})} \\
																								-~\cos{(90^{\circ}-\beta_{P})}	
																			\end{IEEEeqnarraybox*}\right] 																								\\[3pt]	
		f_{PL}	& = & F_{PL} \cdot   \left[\begin{IEEEeqnarraybox*}[\mysmallarraydecl][c]{,c,}
																								+~\sin{(90^{\circ}-\beta_{P})} \cdot \sin{(90^{\circ}-\alpha_{P})} \\	
																								-~\sin{(90^{\circ}-\beta_{P})} \cdot \cos{(90^{\circ}-\alpha_{P})} \\
																								+~\cos{(90^{\circ}-\beta_{P})}			
																			\end{IEEEeqnarraybox*}\right] 
\end{IEEEeqnarray}
\begin{IEEEeqnarray}{rCl}
		f_{VTB}	& = & F_{VTB} \cdot    \left[\begin{IEEEeqnarraybox*}[\mysmallarraydecl][c]{,c,}
																								0 																						\\	
																								0 																						\\
																								1			
																			\end{IEEEeqnarraybox*}\right] 													\\[3pt]	
		f_{VTS}	& = & F_{VTS} \cdot    \left[\begin{IEEEeqnarraybox*}[\mysmallarraydecl][c]{,c,}
																								0 																						\\	
																								0 																						\\
																								1			
																			\end{IEEEeqnarraybox*}\right]		
\end{IEEEeqnarray}
With the distance vectors of the stern propellers (\ensuremath{r_{SbU}}, \ensuremath{r_{SbL}}, \ensuremath{r_{PU}} and \ensuremath{r_{PL}}) and of the vertical thrusters (\ensuremath{r_{VTB}} and \ensuremath{r_{VTS}}), it is possible to calculate the resultant moments. The following addition of the force-moment vectors \ensuremath{\tau_{1}} and \ensuremath{\tau_{2}} forms a vector that contains all influences of the actuators.
\begin{IEEEeqnarray}{rCl}
					\tau				& = &	\left[\begin{IEEEeqnarraybox*}[][c]{,c,}
												f	  \\
												r \times f
									\end{IEEEeqnarraybox*}\right] = \tau_{1} + \tau_{2} 																											\\[3pt]
					\tau_{1}		& = &	\left[\begin{IEEEeqnarraybox*}[\mysmallarraydecl][c]{,c,}
															f_{SbU} + f_{SbL} + f_{PU} + f_{PL} 																													\\
															r_{SbU} \times f_{SbU} + r_{SbL} \times f_{SbL} + r_{PU} \times f_{PU} + r_{PL} \times f_{PL}	
														\end{IEEEeqnarraybox*}\right]																																		\\[3pt]
					\tau_{2}		& = &	\left[\begin{IEEEeqnarraybox*}[\mysmallarraydecl][c]{,c,}
															f_{VTB} + f_{VTS}																																							\\
															r_{VTB} \times f_{VTB} + r_{VTS} \times f_{VTS}								
														\end{IEEEeqnarraybox*}\right]
\end{IEEEeqnarray}
	\subsection{Optimization}
\begin{figure}[!h]
	\centering
	\vspace{0pt}
	\includegraphics[width=0.48\textwidth]{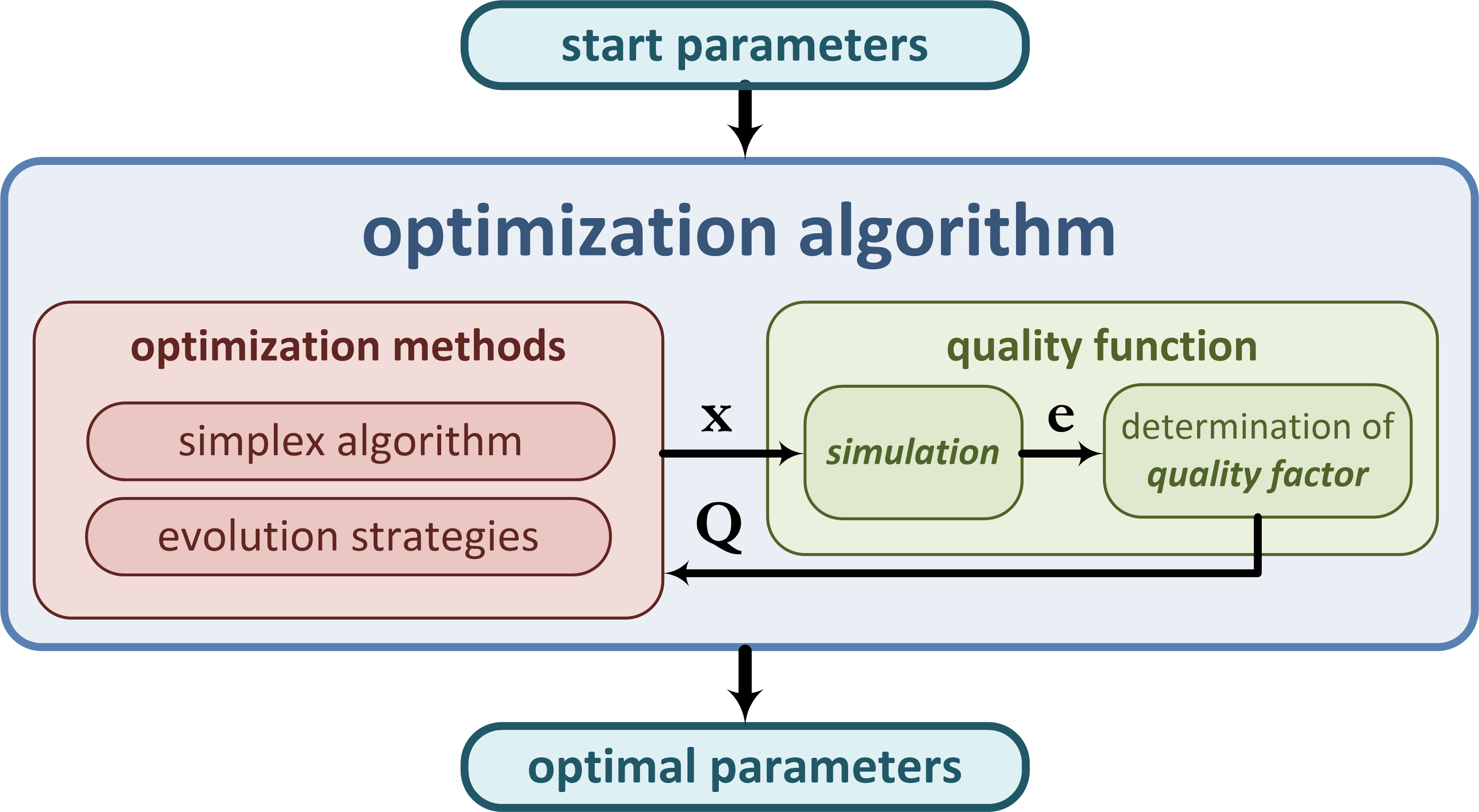}
	\caption{Schematic design of parameter optimization}
	\label{471}
	\vspace{0pt}
\end{figure}

The optimization of the model parameters is carried out by the improvement of associated coefficients \ensuremath{C_{...}} according to the schema shown in Fig. \ref{471}.\\
The start parameters are based on geometric approximations and are placed within the certain limits of optimization. The data recorded in sea trials serve as a reference and provide the control variable vector for the simulation of the model. On the model output, the model error \ensuremath{e} is calculated by comparing the of measured \ensuremath{y(t)} and simulated data \ensuremath{\hat{y}(t)}. The least absolute error criterion determines the quality of the current parameter set \ensuremath{Q}. 
\begin{equation}
		Q = \int |e(t)|~dt = \int |y(t)-\hat{y}(t)|~dt
\end{equation}	
The optimization algorithm seeks the global minimum of the quality by the means of a evolution strategy and uses a simplex algorithm to increase the convergence speed while doing the fine search. The parallel optimization is divided into four stages. Individual force or torque components are considered independently with the vector \ensuremath{V_{const}} (see Fig. \ref{402}) to prevent the mutual coupling effects.
\begin{figure}[!ht]
	\centering
	\vspace{0pt}
	\includegraphics[width=0.48\textwidth]{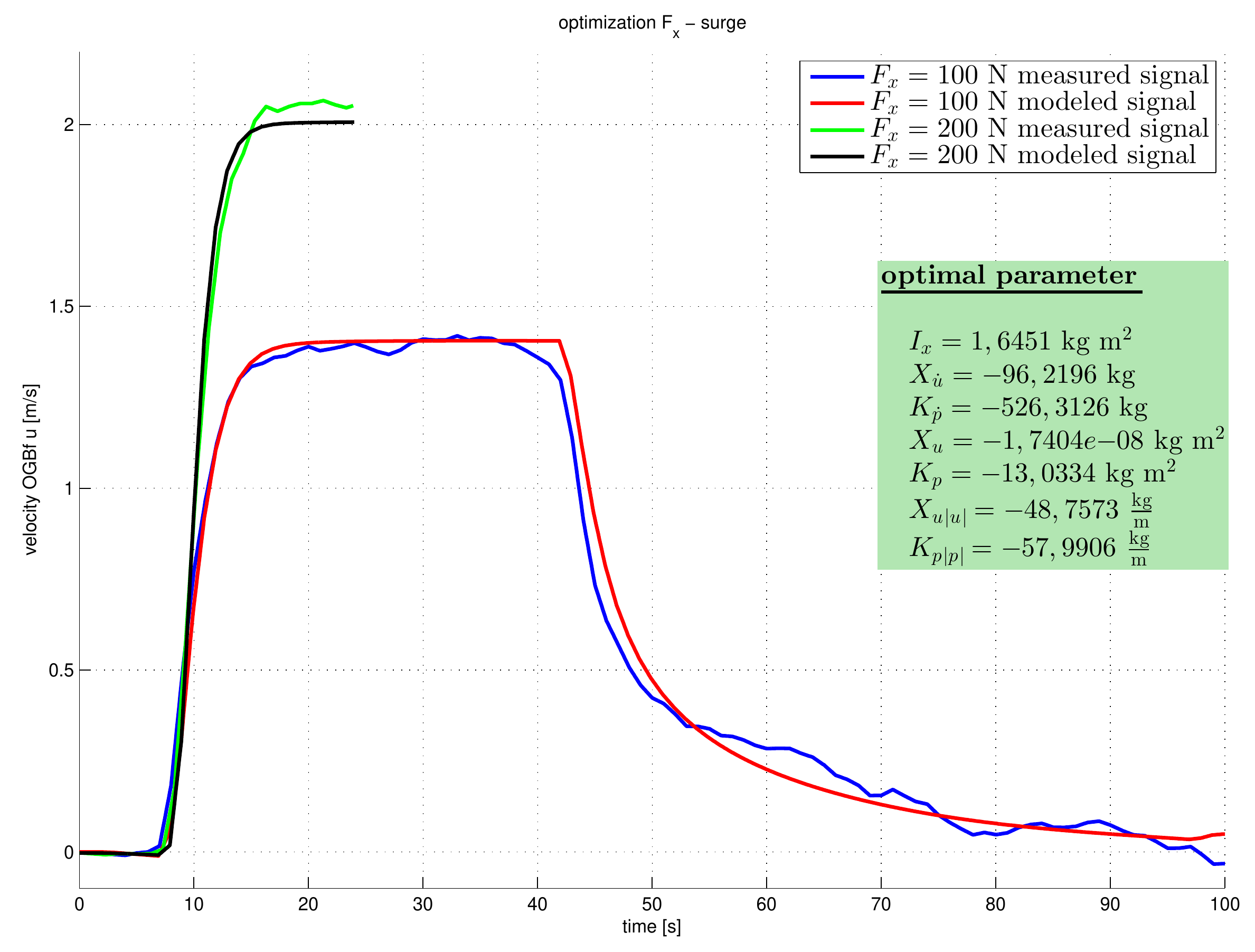}
	\caption{Optimization of the forward speed with stern propellers}
	\label{472}
	\vspace{0pt}
\end{figure}
%
%
\begin{figure}[!ht]
	\centering
	\vspace{0pt}
	\includegraphics[width=0.48\textwidth]{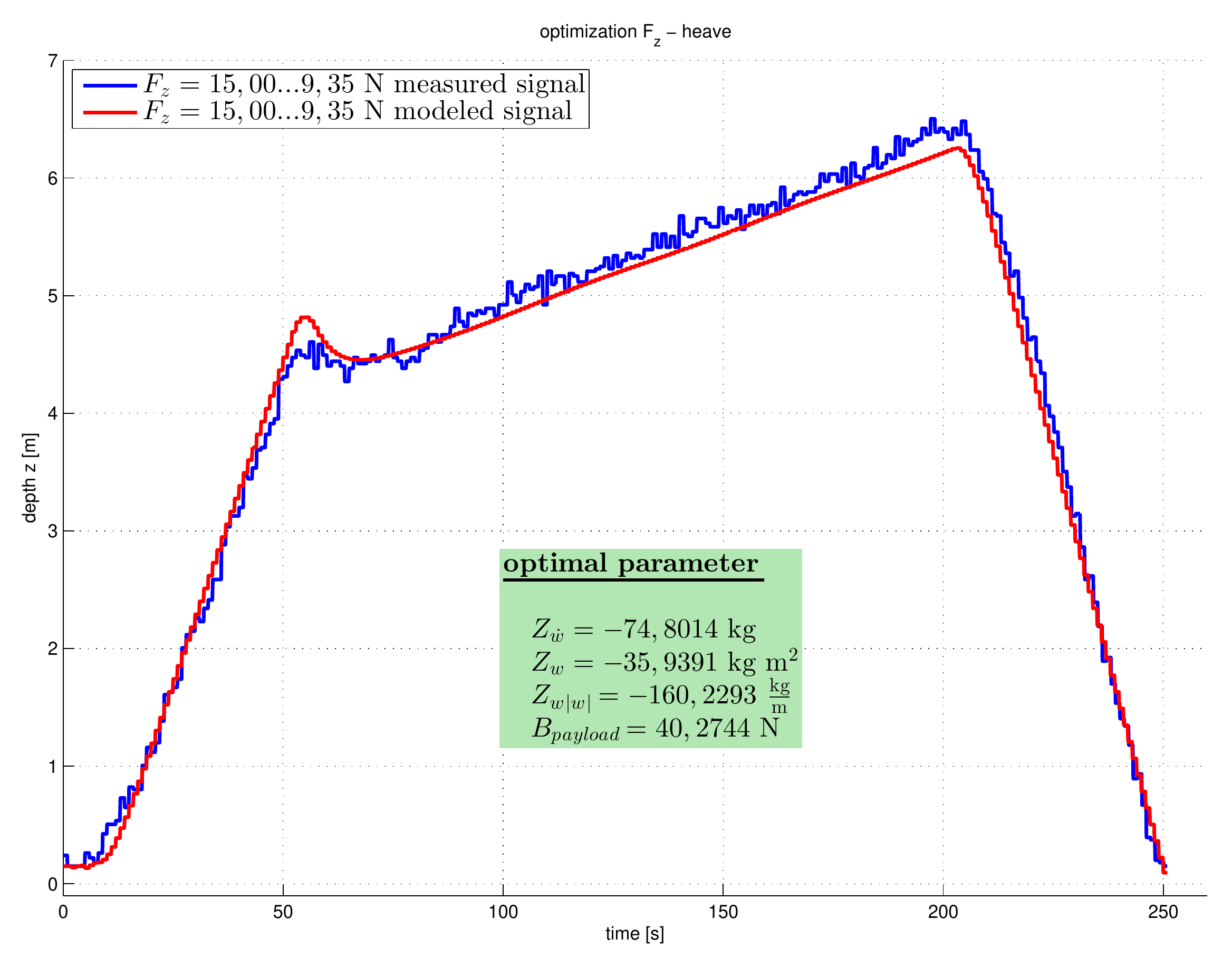}
	\caption{Optimization of the diving with vertical thrusters}
	\label{473}
	\vspace{0pt}
\end{figure}
%
%
\begin{figure}[!ht]
	\centering
	\vspace{0pt}
	\includegraphics[width=0.48\textwidth]{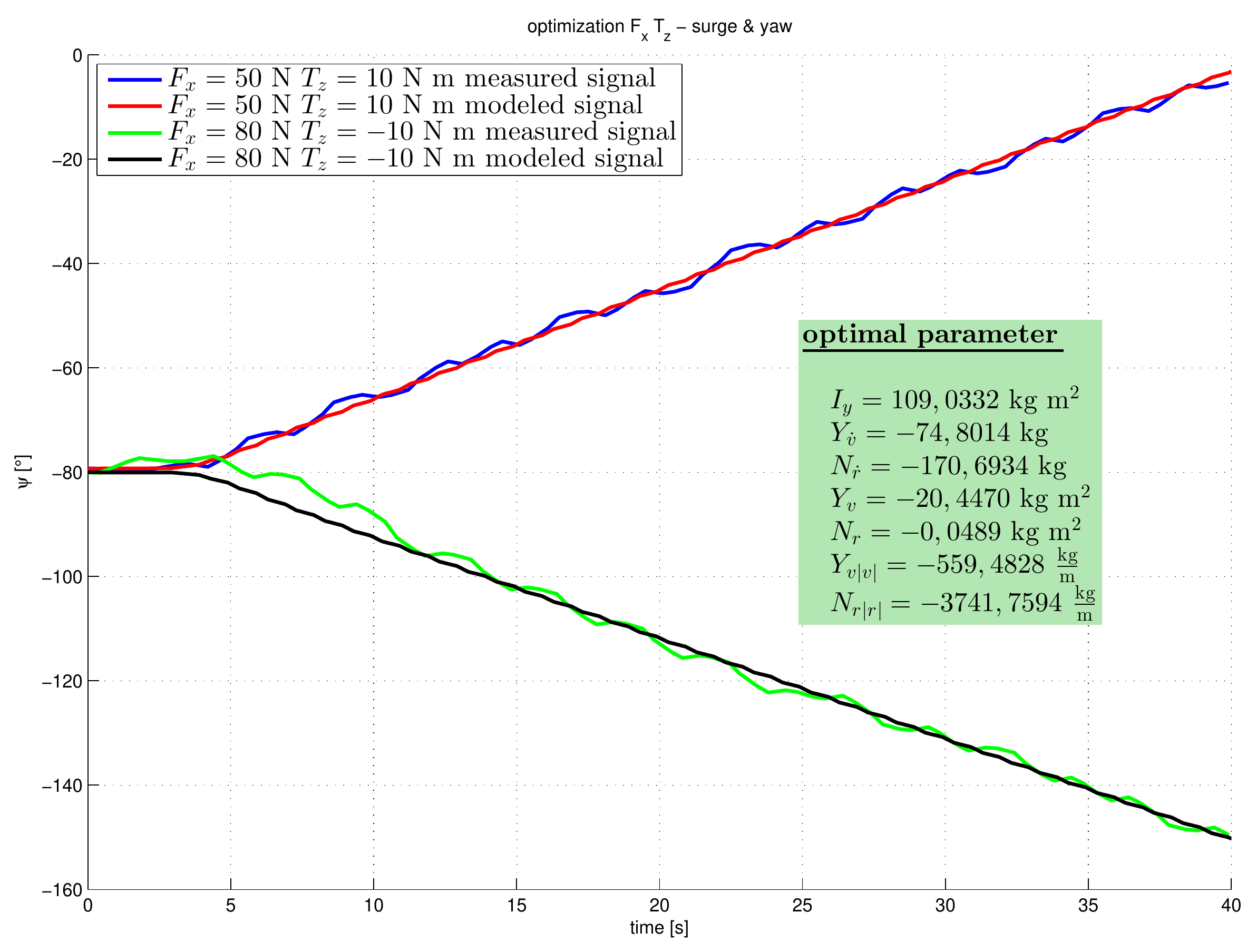}
	\caption{Optimization of the turning with stern propellers}
	\label{474}
	\vspace{-5pt}
\end{figure}
%
%
\begin{figure}[!ht]
	\centering
	\vspace{0pt}
	\includegraphics[width=0.48\textwidth]{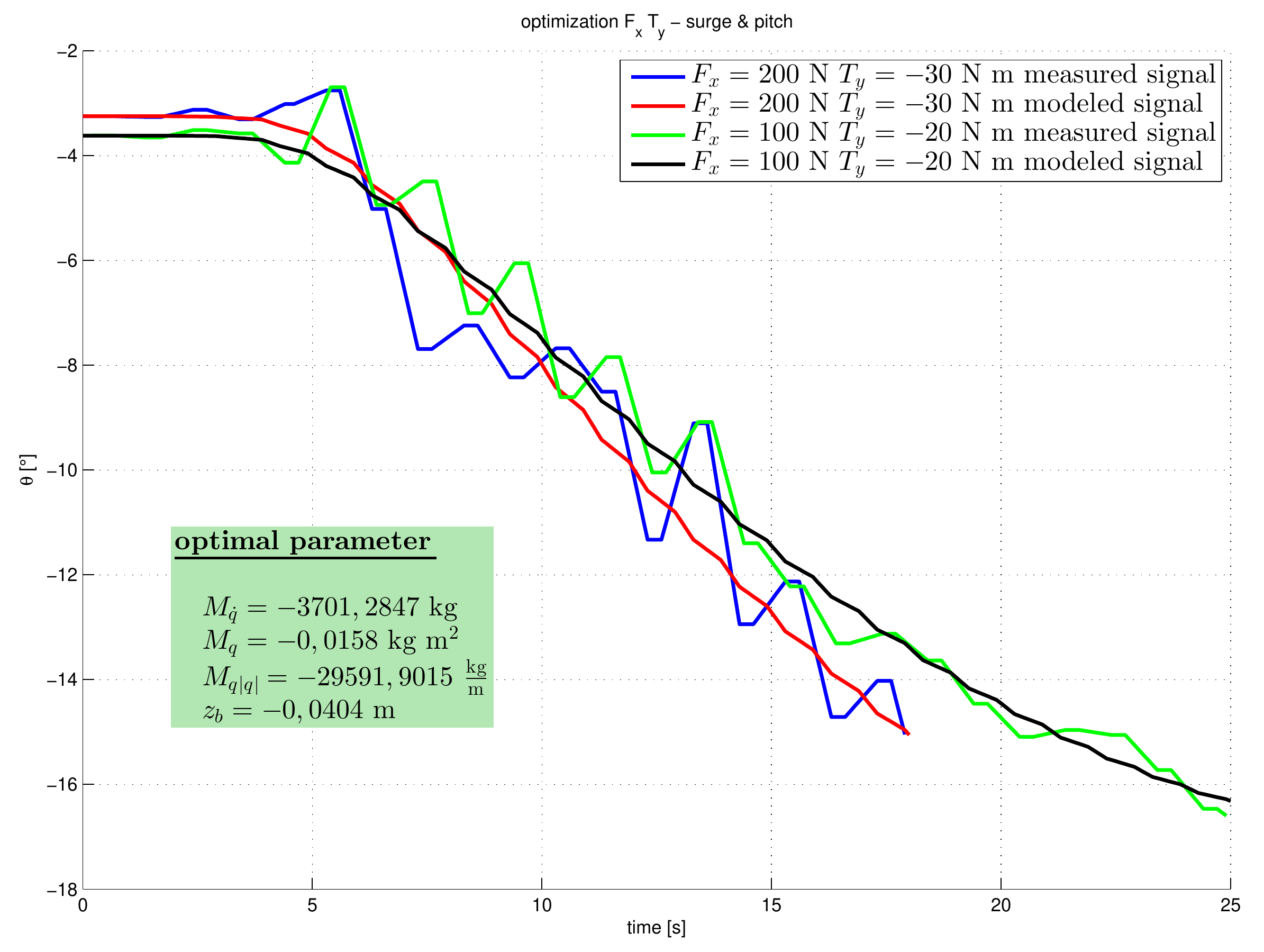}
	\caption{Optimization of the diving with stern propellers}
	\label{475}
	\vspace{-5pt}
\end{figure}

	\section{Controller Design}
\begin{figure}[!ht]
	\centering
	\vspace{0pt}
	\includegraphics[width=0.48\textwidth]{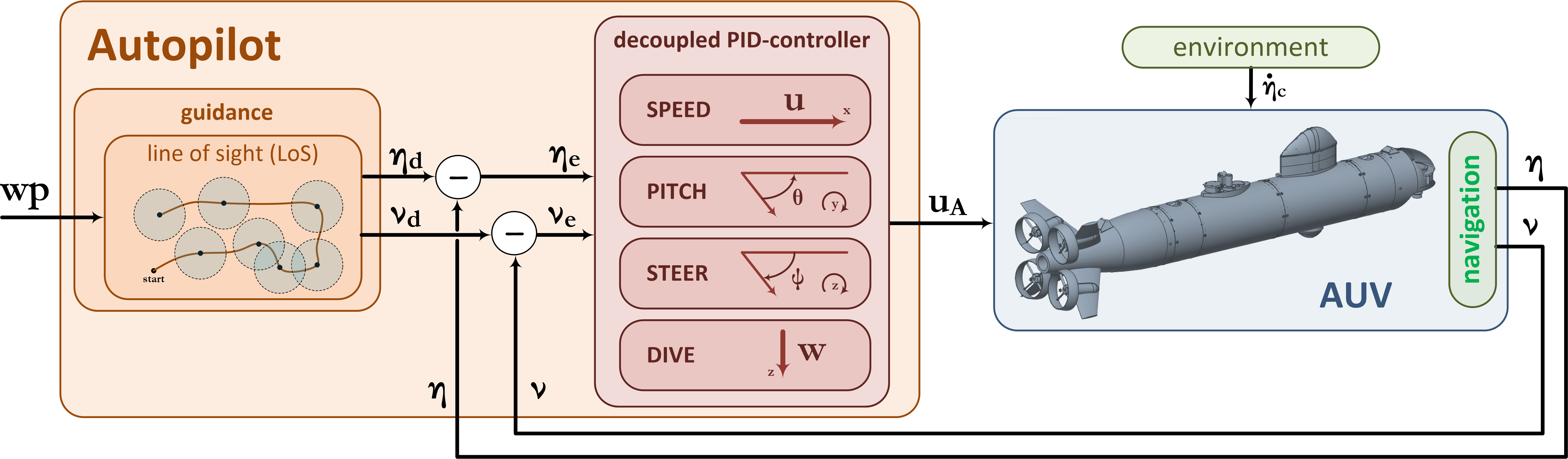}
	\caption{Block diagram of control loop}
	\label{501}
	\vspace{0pt}
\end{figure}

The presented autopilot consists of a waypoint guidance, based on the principle of ``Line of Sight'' (LoS) and four decoupled adaptive PID controllers (Fig \ref{501}).

After calculating the desired values (\ensuremath{\eta_{d}}, \ensuremath{\nu_{d}}) from the specified waypoints \ensuremath{wp}, the control deviation (\ensuremath{\eta_{e}}, \ensuremath{\nu_{e}}) is formed by the measured navigation data (\ensuremath{\eta}, \ensuremath{\nu}). The adaptive PID controller generates the control variable vector \ensuremath{u_{A}}, by means of the inverse actuator matrix \ensuremath{B^{-1}} which serves as the interface to the vehicle.
\begin{IEEEeqnarray}{rCl}
				u_{A}	& = & B^{-1}\left[\tau_{PID}(\nu_{e})+g(\eta)\right]														\\
							& = & B^{-1}\left[J^{T}(\eta_{2})~\tau_{PID}(\eta_{e})+g(\eta)\right]		
\end{IEEEeqnarray}
The additive combination of the restoring forces and moments \ensuremath{g(\eta)} is used as the control variable intrusion, because these influences are known a priori.
	\subsection{Guidance control}
The guidance system generates a ``Line of Sight'' (LoS) vector from the available waypoints \ensuremath{wp_{k} =  \left[x_{k}~~y_{k}~~z_{k} \right]^{T}}, continuously reducing the cross track error \ensuremath{xte}. The 3D-LoS calculates the heading angle \ensuremath{\psi_{d}} and the pitch angle \ensuremath{\theta_{d}}, using the arc tangent function \cite{Karimanzira.2013}.
\begin{IEEEeqnarray}{rCl}
				\theta_{d} 	& = & \text{atan2}~(z_{k}-z(t),x_{k}-x(t))													\\
				\psi_{d} 		& = &	\text{atan2}~(y_{k}-y(t),x_{k}-x(t))		
\end{IEEEeqnarray}
The switchover to the next waypoint occurs when the vehicle enters the sphere of acceptance with the radius \ensuremath{R_{0}}.
\begin{equation}
			\left[x_{k}-x(t)\right]^{2} + \left[(y_{k}-y(t)\right]^{2} + \left[(z_{k}-z(t)\right]^{2} \leq R_{0}^{~2}
\end{equation}
\begin{figure}[!ht]
	\centering
	\vspace{0pt}
	\includegraphics[width=0.40\textwidth]{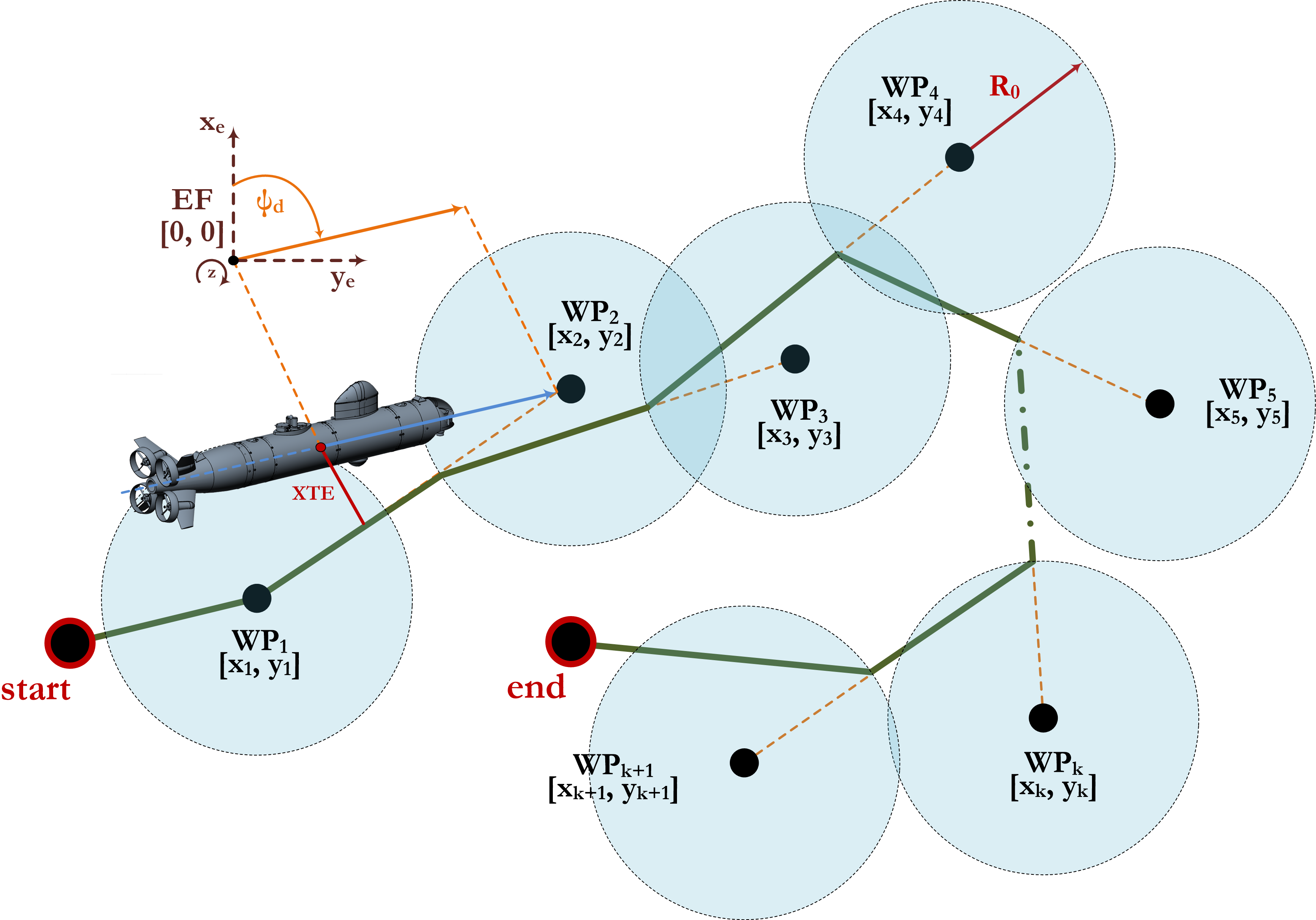}
	\caption{Waypoint guidance with ``Line of Sight''}
	\label{511}
	\vspace{-10pt}
\end{figure}
	\subsection{Adaptive PID-Controller}
The controlled systems offer no constant behaviour, especially while an autopilot operates in closed loop. Due to the dependence of vehicle speed and acceleration it's useful to variably adapt the controller behaviour. A characteristic with six supporting points changes the control parameters (\ensuremath{K_{P}}, \ensuremath{K_{I}}, \ensuremath{K_{D}}) in the velocity range of \ensuremath{-5,0~\rfrac{m}{s} \leq u_{d} \leq 5,0~\rfrac{m}{s}}. 
\begin{IEEEeqnarray}{rCl}
				\tau_{PID} & = 	& K_{P}(u_{d}) \cdot e_{R}(k) 														\IEEEnonumber \\
									 & 		&	+ K_{I}(u_{d}) \cdot \sum^{k-1}_{i=0}e_{R}(i)						\IEEEnonumber \\
									 &		&	- K_{D}(u_{d}) \cdot \left[y_{R}(k) - y_{R}(k-1)\right]
\end{IEEEeqnarray}
The stable desired velocity \ensuremath{u_{d}} is used instead of the fluctuating actual value \ensuremath{u} to avoid instabilities during parameter adaptation. In addition, only the negative controlled variable \ensuremath{y} is used for the calculation of the derivative term of the PID controller in order to prevent an excessive response in sudden changes.

\begin{figure}[!ht]
	\centering
	\vspace{0pt}
	\includegraphics[width=0.48\textwidth]{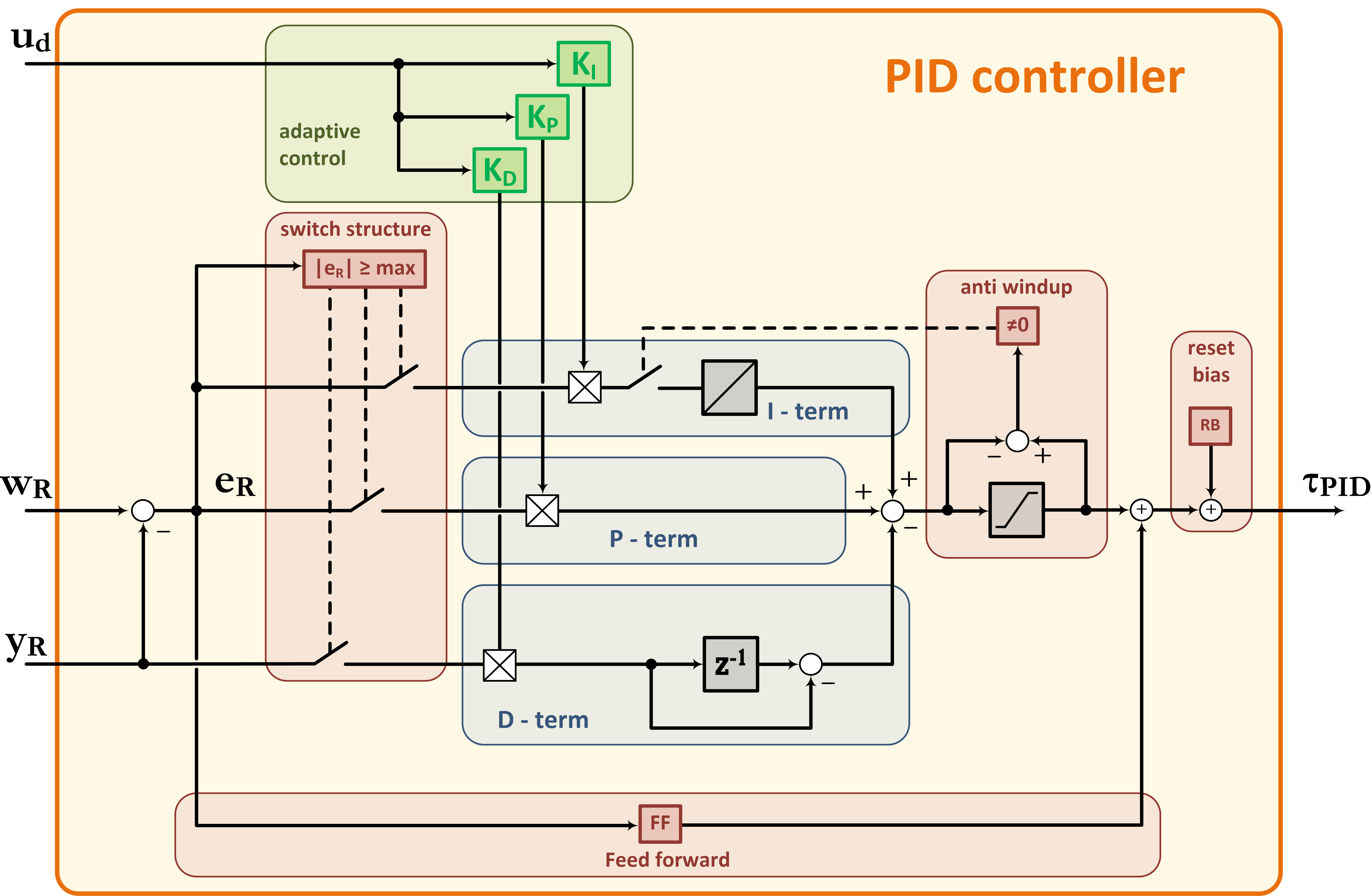}
	\caption{Structure of adaptive PID-controller with additional features}
	\label{521}
	\vspace{0pt}
\end{figure}
For stability and accuracy, controller functions like anti-windup, feed forward, reset bias and switch structure are integrated in the decoupled adaptive controller \cite{Isermann.1988}.
	\subsection{Constraint Criteria Optimization}
The command action of the control loop is provided by the quality criteria. The constraints are specified with the Nonlinear Control Design (NCD) - Blockset for Simulink. By defining the lower \ensuremath{C_{u}(t)} and upper constraint \ensuremath{C_{o}(t)} between a set point response \ensuremath{y(t)} has to run, the controller parameters can be optimized, using the Fuzzy Control Design (FCD) - Toolbox for MATLAB. 

\begin{figure}[!ht]
	\centering
	\vspace{0pt}
	\includegraphics[width=0.40\textwidth]{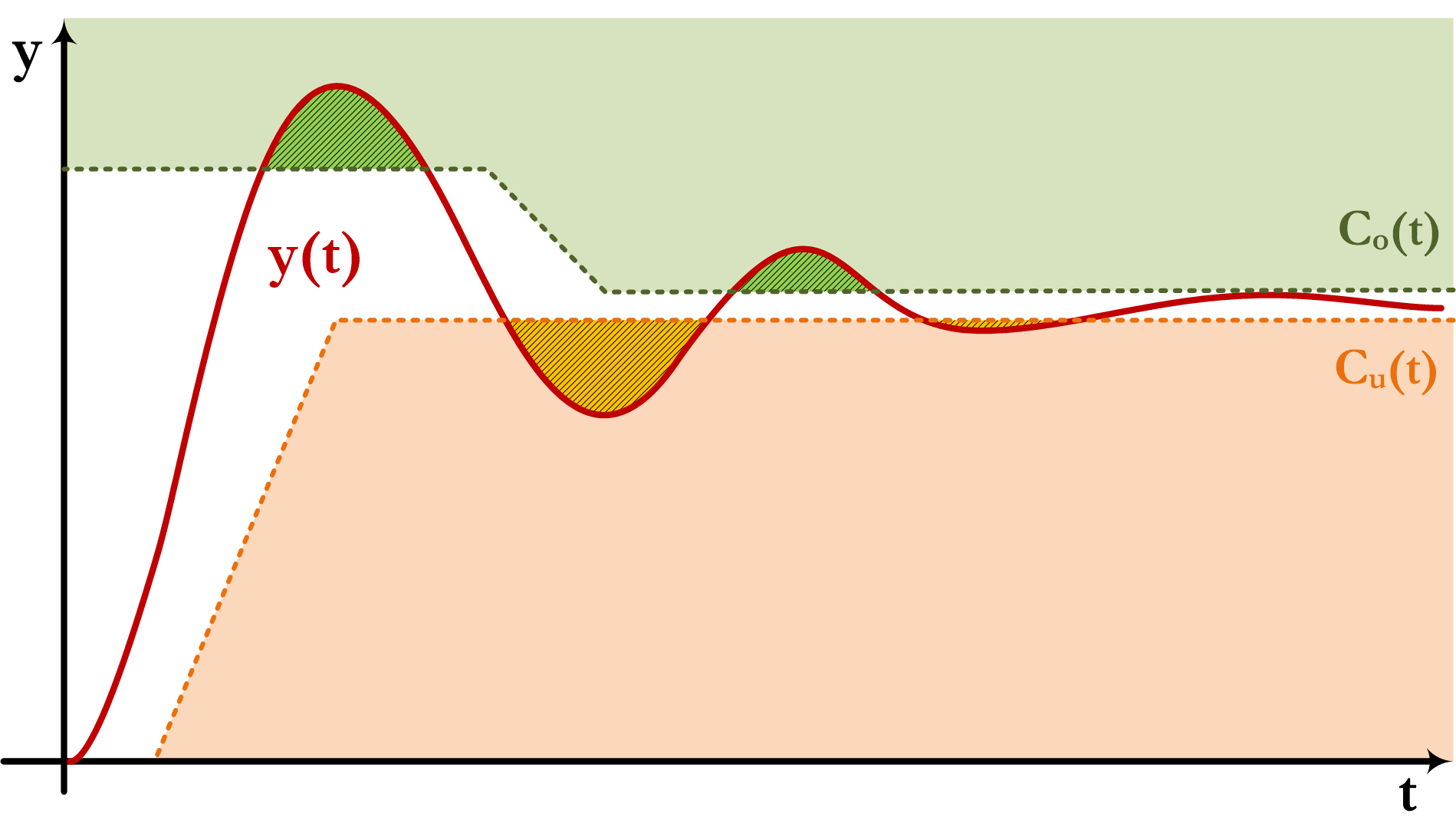}
	\caption{Constraint criteria with error areas}
	\label{551}
	\vspace{0pt}
\end{figure}
\begin{figure}[!ht]
	\centering
	\vspace{0pt}
	\includegraphics[width=0.48\textwidth]{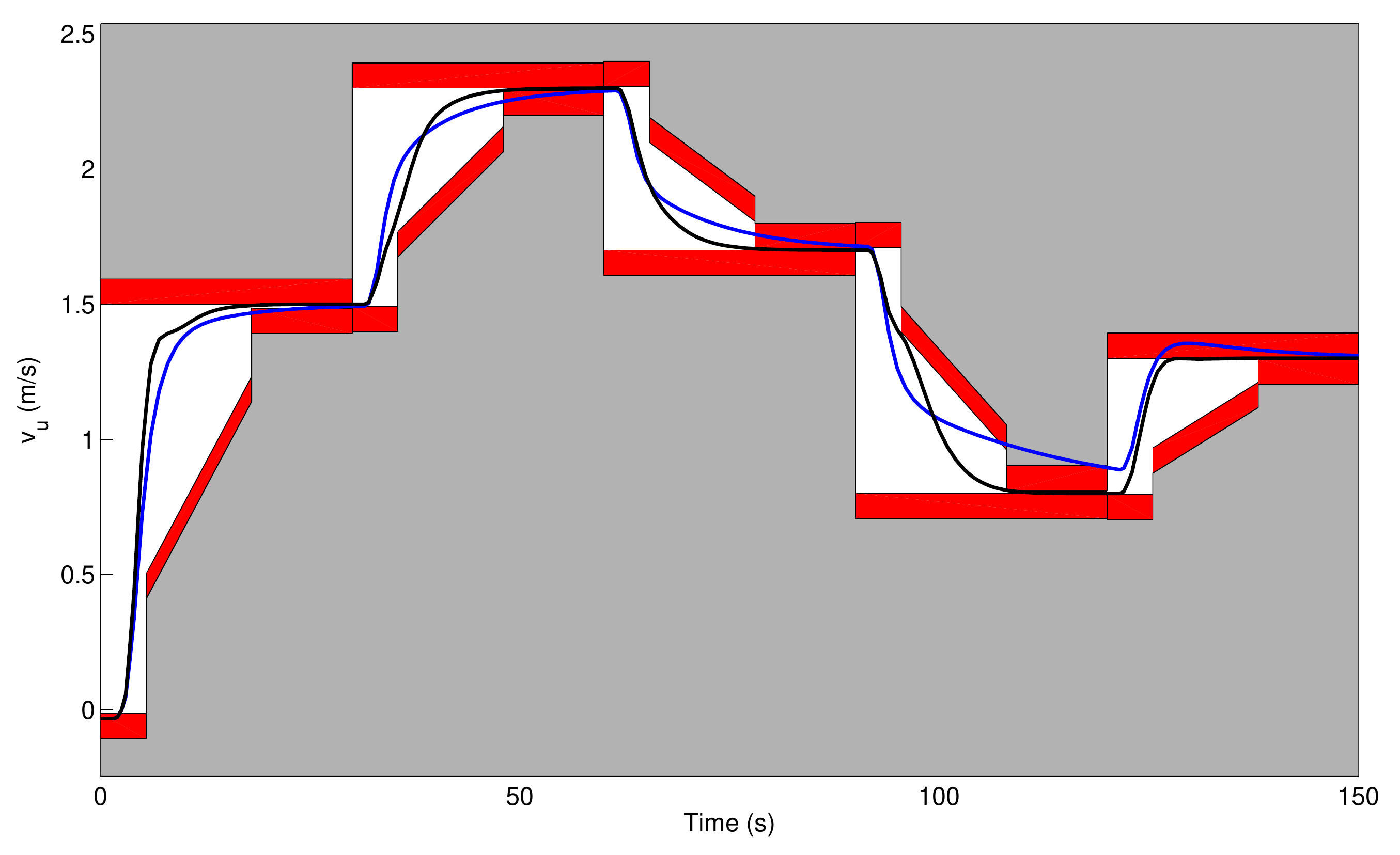}
	\caption{Speed control optimization with constraint criteria}
	\label{581}
	\vspace{0pt}
\end{figure}

In violation of a constraint by the signal trajectory, a flag (\ensuremath{US(t)}, \ensuremath{OS(t)}) is activated. 
\begin{IEEEeqnarray}{rCl}
						US(t)	 & = 	& \left\{\begin{IEEEeqnarraybox}[\mysmallarraydecl][c]{l?s}
																0,&for $C_{u}(t) < y(t)$					\\
																1,&for $C_{u}(t) \geq y(t)$				
																\end{IEEEeqnarraybox}\right.			\\[3pt]
						OS(t)	 & = 	& \left\{\begin{IEEEeqnarraybox}[\mysmallarraydecl][c]{l?s}
																0,&for $C_{o}(t) > y(t)$					\\
																1,&for $C_{o}(t) \leq y(t)$				
																\end{IEEEeqnarraybox}\right.			
\end{IEEEeqnarray}
The quality of the current controller parameter set \ensuremath{Q} is calculated from the weight factor of constraint violation (\ensuremath{WI_{u}}, \ensuremath{WI_{o}}) \cite{Eichhorn.2001}.
\begin{equation}
		Q = \int_{0}^{t} \left( \begin{IEEEeqnarraybox}[\mysmallarraydecl][c]{,l,}
																US(t) \cdot WI_{u} \cdot (C_{u}(t) - y(t)) ~+ ~... \\
																OS(t) \cdot WI_{o} \cdot (y(t) - C_{u}(t))				
															\end{IEEEeqnarraybox}\right)~dt
\end{equation}	
Figure \ref{581} shows the constraint criteria optimized velocity in the x-direction \ensuremath{u}, changing by different set points \ensuremath{u_{d}}.

	\section{Conclusion and Future work}
A concept for a complete modeling of an autonomous underwater vehicle was presented. Based on this optimized model, an autopilot was designed, whose functionality has been verified in sea trials. The procedure can be adapted to other underwater vehicles and provides an applicable template.

The description of the hydrodynamic variables of the model and the designing of the controller parameters could be optimized by conducting extensive sea trials. The current status is based on only two weeks with realistic experiments. 

In addition to the application as a simulator, it's conceivable to use the vehicle model within a Kalman filter to improve the navigation. The robustness and insensitivity to disturbances could be increased by the implementation of more advanced control concepts in the autopilot.


\bibliographystyle{IEEEtran}
\bibliography{IEEEabrv,references}
%


\end{document}